\documentclass[accepted]{uai2025}

\usepackage[american]{babel}

\usepackage{natbib} 
\usepackage{mathtools} 
\usepackage{booktabs} 
\usepackage{tikz} 
\usepackage{graphicx}
\usepackage{amssymb}
\usepackage{bbm}
\usepackage{bm}
\usepackage{enumitem}
\usepackage{multirow}
\usepackage{caption}
\usepackage{subcaption}
\usepackage{wrapfig} 
\usepackage{xspace}
\usepackage{colortbl}
\usepackage{xcolor}
\usepackage{tabularx}

\newif\ifcomments
\commentstrue

\ifcomments
\newcommand{\ba}[1]{{{\textcolor{red}{[B: #1]}}}}
\newcommand{\sj}[1]{{{\textcolor{blue}{[S: #1]}}}}
\newcommand{\yy}[1]{{{\textcolor{Green}{[Yu: #1]}}}}
\newcommand{\hy}[1]{{{\textcolor{purple}{[H: #1]}}}}
\newcommand{\yx}[1]{{{\textcolor{orange}{[X: #1]}}}}
\else
\newcommand{\ba}[1]{}
\newcommand{\sj}[1]{}
\newcommand{\yy}[1]{}
\newcommand{\hy}[1]{}
\newcommand{\yx}[1]{}
\fi

\newcommand{\animals}{\textsc{SpuCoAnimals}\xspace}

\DeclareMathOperator*{\E}{\mathbb{E}}

\DeclareMathOperator*{\argmin}{arg\,min}
\newcommand{\CE}{\text{CE}}

\newcommand{\x}{\pmb{x}}

\newcommand{\w}{\pmb{w}}

\newcommand{\D}{\mathcal{D}}

\newcommand{\spare}{\textsc{Spare}\xspace}
\newcommand{\animal}{\textsc{SpuCoAnimals}\xspace}
\newcommand{\sun}{\textsc{SpuCoSun}\xspace}
\newcommand{\suneasy}{\textsc{SpuCoSun} (Fast)\xspace}
\newcommand{\water}{\textsc{Waterbirds}\xspace}
\newcommand{\celeba}{\textsc{CelebA}\xspace}
\newcommand{\cars}{\textsc{UrbanCars}\xspace}
\begin{document}
\title{Challenges and Opportunities in Improving Worst-Group Generalization in Presence of Spurious Features}

\author[1]{Siddharth Joshi}
\author[1]{Yihao Xue*}
\author[1]{Yu Yang*}
\author[1]{Wenhan Yang*}
\author[1]{Baharan Mirzasoleiman}
\affil[1]{Department of Computer Science, UCLA, Los Angeles, CA, 90024}

\maketitle

\begin{abstract}
Deep neural networks often exploit (\textit{spurious}) features that are present in the majority of examples within a class during training. This leads to \textit{poor worst-group test accuracy} i.e. poor accuracy for minority groups that lack these spurious features. Despite the growing body of recent efforts to address spurious correlations (SC), several challenging settings remain unexplored. In this work, we propose studying methods to mitigate SC in settings with 1) spurious features that are learned more slowly, 2) a larger number of classes and 3) a larger number of groups. We introduce two new datasets, \animals and \sun, to facilitate this study and conduct a systematic benchmarking of \textbf{8} state-of-the-art (SOTA) methods across a total of \textbf{5} vision datasets, training over \textbf{5K} models. Through this, we highlight how existing group inference methods struggle in the presence of spurious features that are learned later in training. Additionally, we demonstrate how all existing methods struggle in settings with more groups and/or classes. Finally, we show the importance of careful model selection (hyperparameter tuning) in extracting optimal performance, especially in the more challenging settings we introduced, and propose more cost-efficient strategies for model selection. Overall, through extensive and systematic experiments, this work uncovers a suite of new challenges and opportunities for improving worst-group generalization in the presence of spurious features.

\end{abstract}
\footnotetext{* = Equal contribution.}
\section{Introduction}
Overparameterized machine learning models can be highly accurate on average, yet consistently fail on atypical or minority groups of the data. These performance disparities across groups can be especially pronounced in the presence of spurious correlations, i.e. features that exist in majority of examples of a class in training data but not in test data. 

There has been a lot of recent efforts to address this problem.
If groups of examples with the spurious features are known, robust training methods such as group balancing \cite{idrissi2022simple} or group distributionally robust optimization (GDRO) \cite{sagawa2020distributionally}, upsample or upweight the minority groups to mitigate the spurious correlations. If group labels are not available, existing methods first infer majority and minority groups and then apply robust training to mitigate the spurious correlations \cite{yang2023identifying,liu2021just,creager2021environment}.

Nevertheless, existing evaluations are limited to very simplistic scenario of binary classification with spurious features that are learned very early during training. In particular, standard benchmark datasets for this problem are Waterbirds \cite{sagawa2020distributionally} and CelebA \cite{liu2015faceattributes}, both with two classes and two groups in each class, where the majority group contains a simple spurious feature that is learned very early during the training. \looseness=-1

\begin{figure*}[t]
    \centering
    \includegraphics[width=\linewidth]{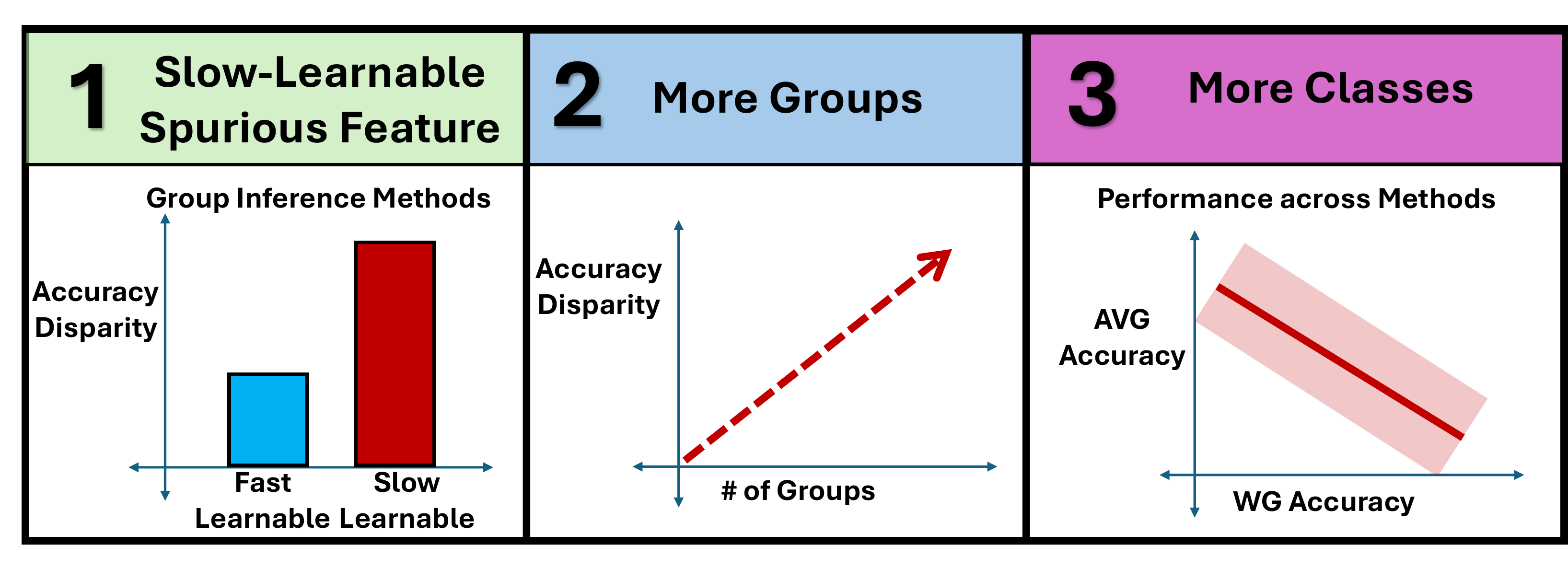} 
    \caption{Larger accuracy disparity is worse. 1) Slow-learnable spurious features are more challenging for group inference methods. 2) More groups are more challenging for all methods 3) More classes make it challenging to mantain high AVG while improving WG accuracy (seen as a strong negative correlation between AVG and WG and a large spread across methods)}
    \label{fig:dataset_promote}
\end{figure*}

In this work, we study the effectiveness of existing methods in previously unexplored settings, i.e., spurious features that are not learned very early during the training and multi-class classification with multiple groups per class. To do so, we introduce two new datasets, namely \sun and \animals, each with four classes and one spurious feature per class, with 16 and 8 groups, respectively. For \sun, we present two versions, \suneasy and \sun, to contrast datasets where the spurious feature is learned early versus later during training. Then, we benchmark 8 state-of-the-art methods on our datasets, in addition to Waterbirds, CelebA, and the more recently introduced UrbanCars \cite{Li_2023_CVPR_Whac_A_Mole} (all with 2 classes).

By training over 5K models, we reveal challenges and opportunities for future research in this direction (summarized in Fig. \ref{fig:dataset_promote}:
\begin{itemize}
    \item When the spurious feature is learned later, group inference methods struggle to infer groups accurately
    \item Existing methods struggle to bridge the gap between worst-group and average accuracy (accuracy disparity) in presence of more groups in the data. 
    \item With the same number of groups, when data has more classes, maintaining high average accuracy while improving worst-group accuracy is more challenging
    \item Model selection is crucial for methods to mitigate SC, especially in the presence of more groups \& classes
    \item Model selection for group inference methods is extremely expensive and can be made $\approx 100\times$ cheaper by directly evaluating the quality of inferred groups 
\end{itemize}
The insights from our study can inspire future research to develop more robust solutions to the critical challenge of improving worst-group accuracy due to spurious features.

\section{Related Work}

\begin{figure}[h]
    \centering
    \includegraphics[width=1\linewidth]{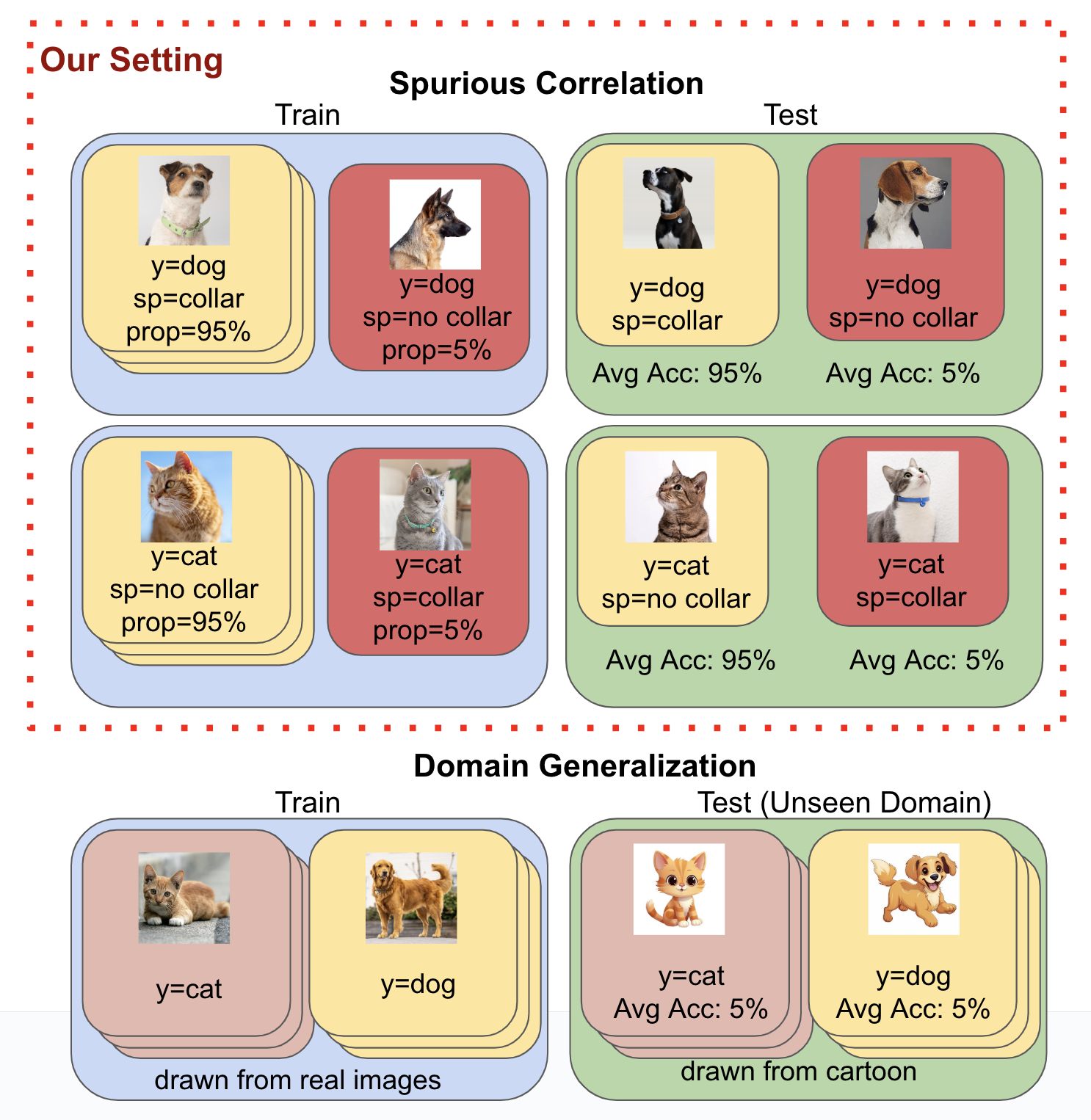}
    \caption{Our Setting (Top): In spurious correlations, certain features (e.g. collar) are spuriously correlated with specific classes (dogs) e.g. Majority of dogs appear with collars and cats without collars, hence collars are spurious correlated with dogs. Bottom: In domain generalization, we train on one domain (e.g., real images) and test on unseen domains (e.g., cartoon images).}
    \label{fig:related}
\end{figure}

\textbf{Spurious correlation.} Spurious correlations occur when a feature is spuriously linked to a class (e.g., dogs often wearing collars, cats typically not), leading to low test accuracy on minority groups (e.g., cats with collars or dogs without). Mitigating models' reliance on spurious correlations can also improve out-of-distribution generalization by encouraging the model to focus on generalizable features (e.g., the true characteristics of a dog) rather than spurious features (e.g., the collar). However, our paper specifically focuses on mitigating spurious correlations in the context of improving worst-group (WG) accuracy, rather than addressing spurious correlations in general or other scenarios like domain generalization \cite{domainbed,zhang2023nico++} (see Fig. \ref{fig:related} for an illustration of the difference between the spurious correlation and domain generalization settings) or other natural distribution shifts \cite{hendrycks2021many,hendrycks2021natural,taori2020measuring}. Improving WG accuracy has been a central objective in a rich body of research on spurious correlations \cite{liu2021just,yang2023identifying,nam2022spread,kirichenko2023last,idrissi2022simple,sagawa2020distributionally,deng2023robust}.

\textbf{Existing Datasets.} 
There are many existing datasets that focus on domain generalization. For example,
DomainBed \cite{domainbed} focuses exclusively on studying domain generalization, and the majority of datasets in WILDS \cite{koh2021wilds} are also designed for this purpose. Other datasets, including NooCh \cite{madras2021identifying}, NICO \cite{he2021towards}, and ImageNet-X \cite{idrissi2022imagenet}, are primarily designed as test sets for evaluating domain generalization, focusing on model performance on unseen domains rather than addressing the spurious correlations setting. Even when spurious correlations are considered, existing datasets often fail to capture challenging settings, such as a large number of groups, multiple classes, or spurious features learned later in training,  which we show are indeed difficult even for current SOTA methods. For example, the datasets in WILDS addressing spurious correlations, such as Waterbirds and CelebA, are limited to only 2 classes and 4 groups, which restricts their effectiveness in evaluating models on more complex spurious correlation challenges. Datasets like bFFHQ \cite{kim2021biaswap} and BAR \cite{nam2020learning} have the same limitations and also exclude certain groups from validation, further reducing their utility for assessing SOTA methods (these rely heavily on model selection using the validation set). Recently, \cite{Li_2023_CVPR_Whac_A_Mole} proposed UrbanCars to study spurious correlations, a dataset with 8 groups, addressing the small \# of groups in previous datasets. 
In contrast to prior work, this paper centers on spurious correlations and introduces datasets with greater complexity: 8-16 groups, 4 classes. Additionally, we also consider settings with fast and slow-learnable spurious features. Our findings reveal significant limitations in current SOTA methods when addressing these complex scenarios.

\textbf{Existing Benchmarks} The recent work by Yang et al. \cite{yang2023identifying} proposes a general framework for subpopulation shifts, including spurious correlations. However, their evaluation is limited, considering only three methods in spurious correlation settings, and their datasets (i.e., WILDS and CelebA) are less challenging. In comparison, our benchmark provides a comprehensive analysis of \textbf{8 SOTA methods} across \textbf{5 vision datasets}, specifically designed to include more groups, more classes, and more challenging spurious features. This extensive evaluation highlights the challenges that remain in mitigating spurious correlations.

\section{Background}

Let $\mathcal{D}=\{(\x_i,y_i)\}_{i=1}^n$ be the training data with input features $\x_i \in \mathbb{R}^d$ and labels $y_i \in [\mathcal{C}]$.  

Machine learning models are trained by minimizing an empirical risk function (ERM) using (stochastic) gradient descent. The goal of ERM is to minimize the average error on the entire training data $\mathcal{D}$: 
\begin{align}\label{eq:erm}
    \w^*\in \argmin_{\w}
    \E_{(\x_i,y_i)\in\D}[l(f(\w,\x_i),y_i)],
\end{align}
where $\w$ is the model parameter, and $f(\w,\x_i)$ and $l(f(\w,\x_i),y_i)$ are the output of the network and the loss associated with a training example $(\x_i, y_i)$, respectively.

\textbf{Spurious correlations and majority groups. }
Let $\mathcal{S}$ be a set of \textit{spurious} features that can be shared between classes. 
Examples in the dataset can be partitioned into different groups $g_{c, s}$, based on the combinations of their label and spurious feature $(c, s)$, where $c\in [\mathcal{C}], s \in \mathcal{S}$. If a spurious feature $s$ appears in majority of training examples of a class $c$, the group $g_{c, s}$ is called the \textit{majority} group in class $c$. Other examples in class $c$ are called \textit{minority}. The ratio of majority group size to minority group size determines the degree of \textit{spurious correlation} between spurious feature $s$ and label $c$. Spurious features have a high correlation with a class in training data, resulting in poor test accuracy when encountering examples without the spurious features.

\textbf{Fast- vs slow-learnable spurious feature.} Fast-learnable features are those that learned early in training. These are typically features with little variation (e.g. only white balls), are simpler to learn and hence are typically learned by the network in early training epochs \cite{nguyen2024make}. In contrast, slow-learnable features are those learned later in training. These have higher variation (e.g. balls that can be any color) and consequently are more challenging to learn. Fast or slow-learnable spurious features, thus, refer to features that are spurious and learned either early or late in training. \looseness=-1 

\textbf{Worst-group accuracy}: Worst-group (WG) test accuracy refers to the lowest accuracy, at test time, across all groups $\mathcal{G}=\bigcup_{(c,s)} g_{c,s}$ \cite{yang2023identifying, sagawa2020distributionally, creager2021environment, liu2021just}. Formally, WG test accuracy is defined as:
\[
\text{WG} := \min_{g \in \mathcal{G}} \mathbb{E}_{(\mathbf{x}_i, y_i) \in g} \left[ y_i = \hat{y}(\mathbf{w}, \mathbf{x}_i) \right],
\]
where $\hat{y}(\mathbf{w}, \mathbf{x}_i)$ is the label predicted by the model.

\textbf{Average accuracy}: Average (AVG) test accuracy refers to a weighted average of test accuracy across groups, where the weight for each group is the fraction of examples in the training set from that group \cite{yang2023identifying, sagawa2020distributionally, creager2021environment, liu2021just}. This quantifies the model performance on test data that i.i.d. to the training data. Formally, AVG test accuracy is defined as:
\[
\text{AVG} := \sum_{g \in \mathcal{G}} \frac{|g|}{|D|} \mathbb{E}_{(\mathbf{x}_i, y_i) \in g} \left[ y_i = \hat{y}(\mathbf{w}, \mathbf{x}_i) \right],
\]
where $\hat{y}(\mathbf{w}, \mathbf{x}_i)$ is the label predicted by the model.

\textbf{Accuracy Disparity}: Accuracy disparity typically refers to the phenomenon where different groups in the data encounter different accuracies \cite{chi2021understanding, barocas2023fairness}. Here, we quantify this disparity using the gap between WG and AVG accuracy. Formally, we define accuracy disparity as:
\[
\text{DISPARITY} := \text{WG} - \text{AVG}.
\]

\section{Experiment Details} \label{sec:setup}

\textbf{Methods.} We consider \textbf{8} SOTA methods aimed at improving worst-group accuracy affected by spurious correlations. We believe this is the first work to comprehensively evaluate SOTA methods specifically for this spurious correlations problem. Concretely, we cover the following algorithms in three categories: \textit{vanilla:} \textbf{ERM} \cite{vapnik}; (1) \textit{group inference methods:} \textbf{EIIL} \cite{creager2021environment}, \textbf{JTT} \cite{liu2021just}, \textbf{SPARE} \cite{yang2023identifying}; (2) \textit{validation group-known methods:} \textbf{SSA} \cite{nam2022spread}, \textbf{DFR} \cite{kirichenko2023last}; (3) \textit{group-known robust methods:} \textbf{GB} \cite{idrissi2022simple}, \textbf{GDRO} \cite{sagawa2020distributionally}, \textbf{PDE} \cite{deng2023robust}. If group labels are available, group-known robust optimization (GDRO) \cite{sagawa2020distributionally} or upsampling the minority groups \cite{liu2021just,yang2023identifying} can robustly train the model against spurious correlations, improving worst-group accuracy. Otherwise, group inference methods first infer groups within the training data based on a reference model trained with ERM, and then retrain the model using GDRO or sampling techniques with these inferred group labels. Validation group-known methods directly utilize the validation data to either infer the training groups or train the ERM model robustly. Appendix \ref{appendix:methods} provides a detailed descriptions for each method.

\textbf{Model selection via validation set.} Model selection (also known as hyperparameter tuning) is critical for all SOTA methods, much like in domain generalization approaches \cite{domainbed}. Poor model selection, even from hyperparameter combinations from carefully chosen ranges, can lead to significant drops (over 70\%) in worst-group test accuracy (see Sec. \ref{sec:model_sel}). To ensure fair evaluation, we extensively tune hyperparameters for these methods based on group-labeled validation data. Following \cite{gulrajani2020search, yang2023change}, we randomly select 16 hyperparameter combinations from the grid, choose the top 3 combinations based on worst-group accuracy on the validation set, and re-run each with 3 different random seeds to report the best average result and its corresponding standard deviation. This procedure ensures that our comparison is best-versus-best, with optimized hyperparameters for all algorithms. In total, we trained over 5K models.

\section{Fast- \& Slow-Learnable Spurious Features}

We now study the effectiveness of existing methods in addressing fast and slow-learnable spurious features. \citet{yang2023identifying} demonstrated that models learn spurious correlations since spurious features are learned faster than true class features. Since a majority of each class can be classified correctly at training time using the spurious features, models do not learn the true class features and perform poorly on minority groups that lack the spurious features. To further investigate the impact of how quickly spurious features are learned on existing methods, we conduct a controlled study comparing fast and slow-learnable spurious features (both features are still learned faster than the true class feature so that models still learn the spurious correlations). Specifically, we introduce a new dataset \sun, with a slow-learnable spurious feature and an alternate version, \suneasy, which has a fast-learnable spurious feature. For both versions, since the spurious features are learned faster than the true class features, ERM exhibits very poor WG accuracy. Interestingly, our analysis reveals that spurious correlations arising from slow learnable spurious features are more challenging to mitigate for group inference methods. This highlights an open challenge for future group inference methods to address.\looseness=-1


\begin{table}[h]
\caption{Effect of Spurious Feature's Difficulty: Larger accuracy disparity for group inference methods with slow-learnable spurious feature \looseness=-1}
\label{tab:spu_diff_tab}
\vspace{-3mm}
\begin{center}
\resizebox{0.95\linewidth}{!}{%
\begin{tabular}{|l|cc|cc|}
\toprule
    & \multicolumn{2}{c|}{\cellcolor{red!10}\sun}& \multicolumn{2}{c|}{\cellcolor{orange!10}\suneasy}\\
 & \cellcolor{red!10}WG & \cellcolor{red!10}AVG & \cellcolor{orange!10}WG & \cellcolor{orange!10}AVG \\\midrule\midrule
 ERM & \cellcolor{red!10}$29.3_{\pm 1.4}$ & \cellcolor{red!10}$96.0_{\pm 0.2}$ & \cellcolor{orange!10}$21.9_{\pm 3.51}$ & \cellcolor{orange!10}$96.7_{\pm 0.1}$\\\midrule\midrule
 JTT & \cellcolor{red!10}$55.8_{\pm 1.1}$ & \cellcolor{red!10}$88.8_{\pm 0.6}$ & \cellcolor{orange!10}$63.5_{\pm 0.3}$ & \cellcolor{orange!10}$85.1_{\pm 0.4}$\\
 \spare & \cellcolor{red!10}$62.2_{\pm 1.4}$ & \cellcolor{red!10}$80.3_{\pm 1.8}$ & \cellcolor{orange!10}$66.3_{\pm 1.6}$ & \cellcolor{orange!10}$79.1_{\pm 1.4}$\\
 EIIL & \cellcolor{red!10}$65.9_{\pm 1.3}$ & \cellcolor{red!10}$76.1_{\pm 3.1}$ & \cellcolor{orange!10}$64.4_{\pm 3.2}$ & \cellcolor{orange!10}$79.0_{\pm 4.2}$\\
 \midrule\midrule
 SSA & \cellcolor{red!10}$68.7_{\pm 0.8}$ & \cellcolor{red!10}$83.9_{\pm 0.8}$ & \cellcolor{orange!10}$66.0_{\pm 0.6}$ & \cellcolor{orange!10}$79.4_{\pm 4.9}$\\ 
 $\text{DFR}^{\text{Val}}_{\text{Tr}}$ & \cellcolor{red!10}$67.3_{\pm 1.6}$ & \cellcolor{red!10}$79.9_{\pm 2.6}$ & \cellcolor{orange!10}$68.2_{\pm 0.7}$ & \cellcolor{orange!10}$80.1_{\pm 1.5}$\\
 \midrule\midrule
 GB & \cellcolor{red!10}$69.9_{\pm 0.6}$ & \cellcolor{red!10}$79.1_{\pm 0.2}$ & \cellcolor{orange!10}$64.4_{\pm 1.5}$ & \cellcolor{orange!10}$78.1_{\pm 0.6}$\\
 GDRO & \cellcolor{red!10}$65.5_{\pm 2.3}$ & \cellcolor{red!10}$78.2_{\pm 0.8}$ & \cellcolor{orange!10}$64.7_{\pm 1.8}$ & \cellcolor{orange!10}$77.4_{\pm 4.8}$\\
 PDE & \cellcolor{red!10}$67.6_{\pm 0.8}$ & \cellcolor{red!10}$77.7_{\pm 1.2}$ & \cellcolor{orange!10}$66.0_{\pm 2.7}$ & \cellcolor{orange!10}$77.4_{\pm 0.3}$\\ \bottomrule
  \multicolumn{1}{c}{} & \multicolumn{2}{c}{\textcolor{red}{\textbf{Slow-Learnable}}} & \multicolumn{2}{c}{\textcolor{orange}{\textbf{Fast-Learnable}}}\\
\end{tabular}%
}
\end{center}
\vspace{-5mm}
\end{table}

\subsection{Datasets}\label{sec:spu_diff:data}

\begin{figure*}[!h]
    \centering
    \begin{subfigure}[t]{0.615\textwidth}
        \centering
        \includegraphics[width=\linewidth]{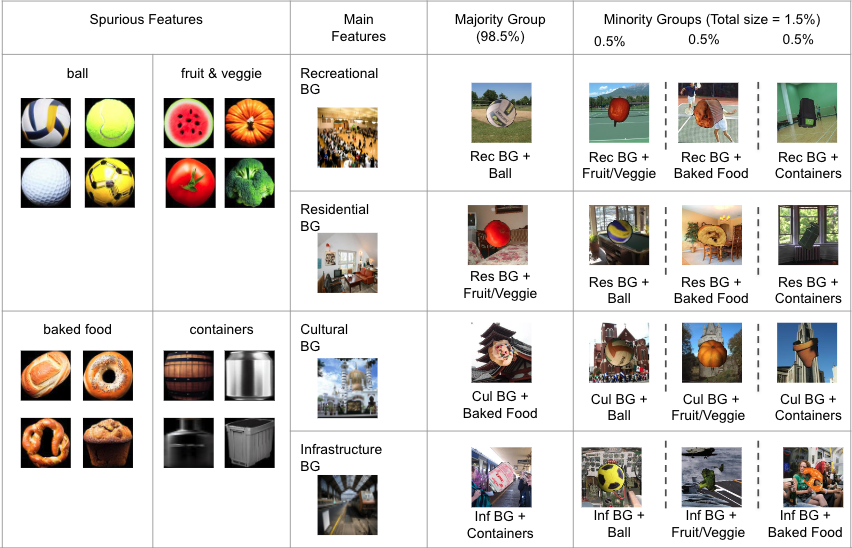}
        \caption{\sun}
        \label{fig:spucosun_data}
    \end{subfigure}
    \hfill
    \begin{subfigure}[t]{0.37\textwidth}
        \centering
        \includegraphics[width=\linewidth]{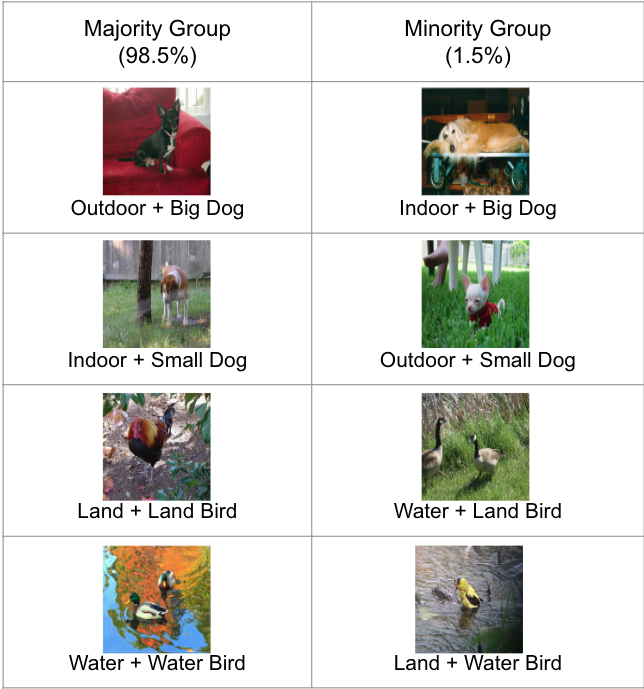}
        \caption{\animal}
        \label{fig:spucoanimals_data}
    \end{subfigure}
    \caption{Construction and examples of images from our \sun and \animals datasets}
    \label{fig:data}
\end{figure*}

\textbf{\sun (16 groups, 4 classes)}. \sun is a dataset with 4 classes and 16 groups. The classes consist of 4 different types of backgrounds selected from the SUN397 dataset \cite{sun}, namely \{\texttt{recreational}, \texttt{residential}, \texttt{cultural}, \texttt{infrastructure}\}. For each class, we generate spurious features as co-occurring objects, corresponding to categories from OpenImagesV7 \cite{OpenImages}. These co-occurring objects are generated using a text-to-image diffusion model \cite{Rombach_2022_CVPR}. The co-occurring objects for the aforementioned 4 classes are \textit{sports equipment}, \textit{fruits \& vegetables}, \textit{baked goods}, and \textit{containers}, respectively. We use \sun and \suneasy to refer to the versions with slow and fast learnable spurious features, respectively. For the slow-learnable spurious features, we sample three subcategories (from \cite{OpenImages}) to generate the spurious feature. Specifically, for sports equipment, we select basketball, golf ball, and tennis ball; for fruits \& vegetables, we select pumpkin, watermelon, and broccoli; for baked goods, we select muffin, bagel, and pretzel; and for containers, we select waste container, can, and barrel. For the fast-learnable spurious features, we use only a single subcategory per spurious feature: sports equipment: golf ball, fruits \& vegetables: watermelon, baked goods: bagel, and containers: barrel. There are 16 groups, specified by all the possible combinations of \textit{class} (background) and \textit{spurious} (co-occurring object) features. The majority groups are: 1) Recreational (BG) + Sports Equipment (CO Obj), 2) Residential (BG) + Fruits \& Vegetables (CO Obj), 3) Cultural (BG) + Baked Goods (Co Obj) and 4) Infrastructure (BG) + Containers (Co Obj). The training set contains 28,092 examples per class, with the majority group comprising 98.5\% of the examples, while each of the three minority groups accounts for 0.5\%. The validation set has 1612 examples with balanced groups. The test set is group-balanced and has 4012 examples. Corresponding images in the two versions, \sun and \suneasy, are identical in all regards, except the spurious feature. Images from \sun are shown in Fig. \ref{fig:spucosun_data}. Figure \ref{fig:spucosun_easiness_evidence} shows that the spurious feature of \suneasy is learned faster than the spurious feature of \sun. \looseness=-1

\begin{figure}[h]
    \centering\includegraphics[width=0.9\linewidth]{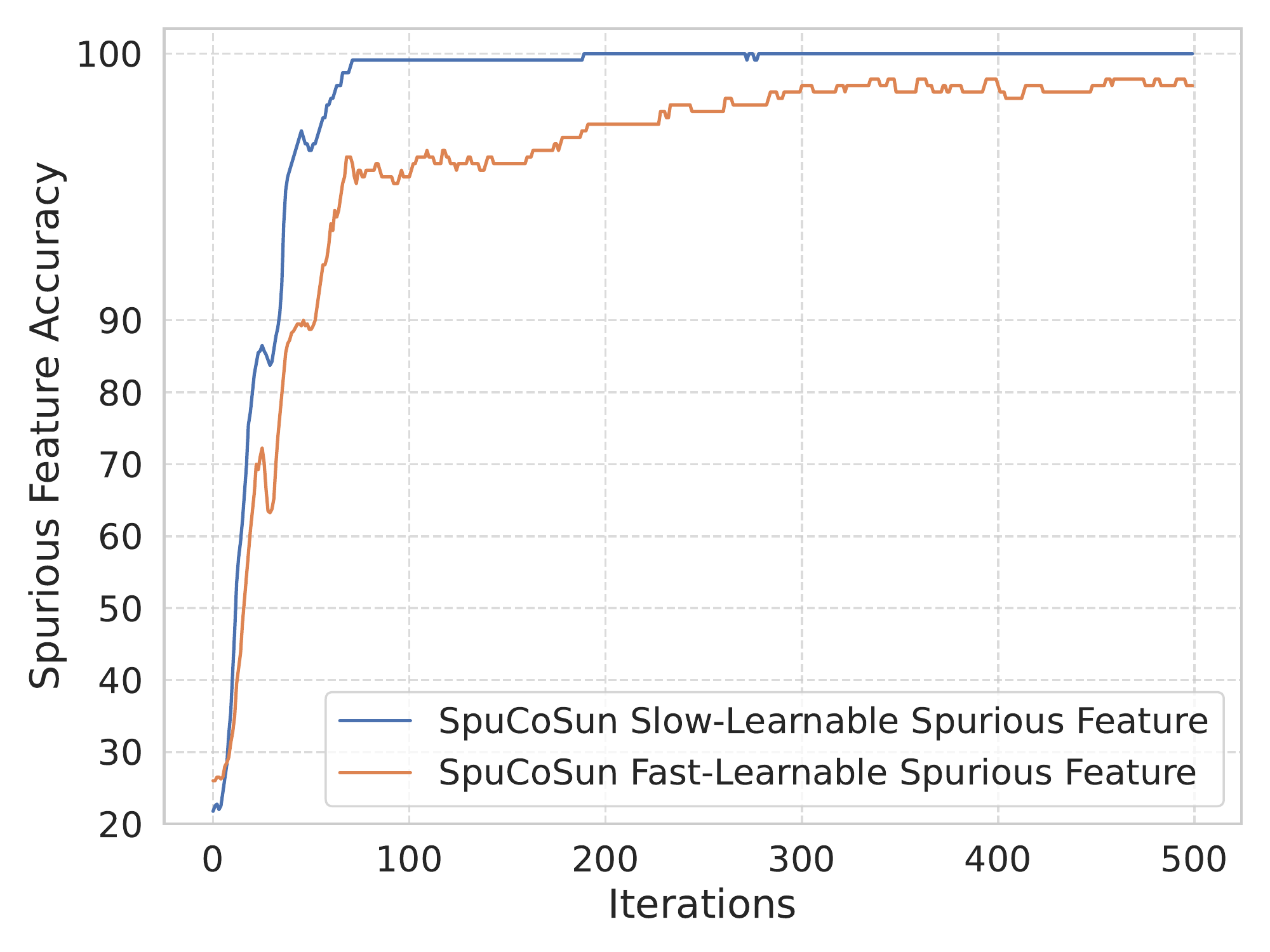}
    \caption{Comparing Average Accuracies of Majority and Minority Groups on \textsc{\suneasy} v/s \textsc{\sun}}
    \label{fig:spucosun_easiness_evidence}
\end{figure}

\begin{figure}
    \centering
    \includegraphics[width=0.9\linewidth]{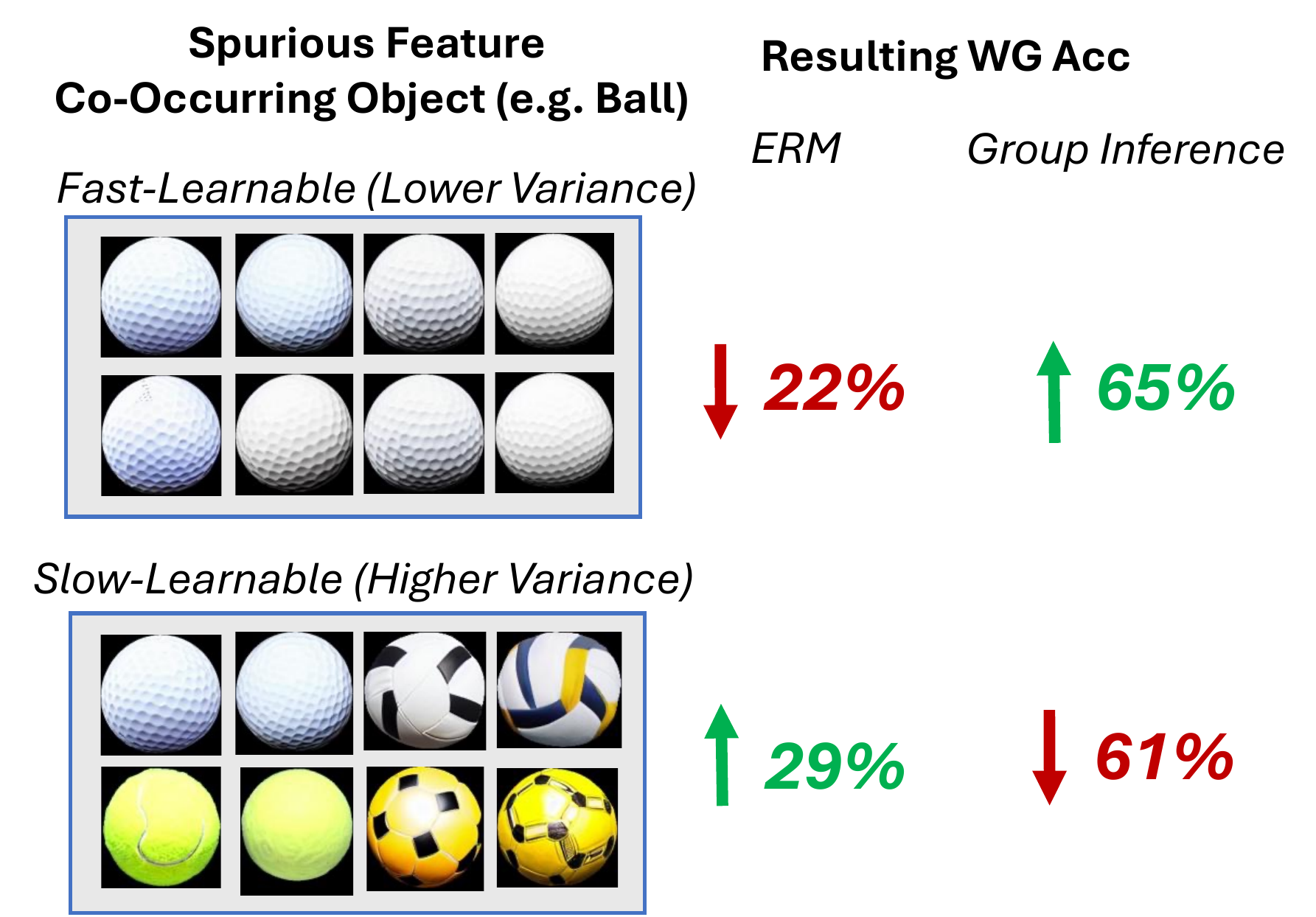}
    \caption{(Left) \suneasy v/s \sun spurious feature, (Right) Both lead to poor WG accuracy with ERM, but group inference methods lag behind on improving WG Accuracy for slow-learnable spurious features}
    \label{fig:spu_diff_construct}\vspace{-2mm}
\end{figure}

\subsection{Results and Analysis}

Table \ref{tab:spu_diff_tab} compares current SOTA methods across \sun and \suneasy, showing how different methods are affected by how quickly the spurious features are learned. Exact hyperparamters and experimental details appear in Appendix \ref{appendix:hparams_main}.

\vspace{-3mm}
\paragraph{Group-inference Methods Struggle in Presence of Slow-learnable Spurious Features.}
Fig. \ref{fig:spu_diff_construct} compares the WG accuracy of ERM with the WG accuracy averaged across all Group Inference methods. While fast-learnable spurious features harm the WG accuracy of ERM more than slow-learnable spurious features, they are easier to address for Group Inference methods. Remarkably, when the spurious feature is learned quickly, the best Group Inference method (\spare) performs comparably to the best method with group labels, PDE, in terms of both worst-group and average accuracy, as seen on \suneasy. However, with slow-learnable spurious features, the performance of \spare and JTT is significantly negatively impacted. Even EIIL, which appears as the best Group Inference method under more slowly learned spurious features and is relatively robust across the two datasets, falls far behind the performance of group-known methods like PDE on \sun.

\begin{figure}[h]
    \centering
    \includegraphics[width=0.8\linewidth]{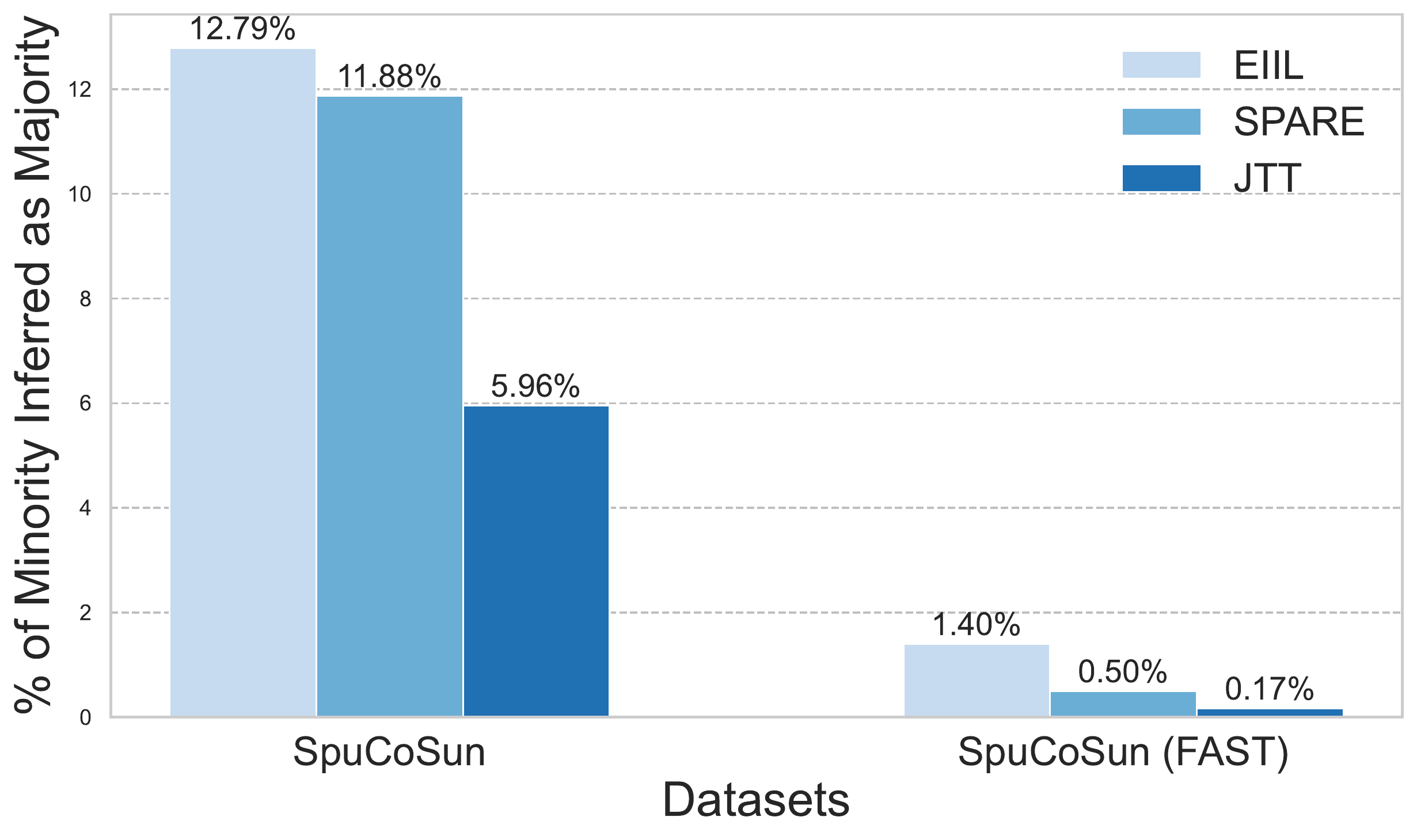}
    \caption{Large Fraction of Minority Inferred as Majority during Group Inference with Slow-Learnable Spurious Features}
    \label{fig:group_infer_slow}
\end{figure}

\textbf{Inferred Group Memberhsip is Less Accurate in the Presence of Slow-learnable Spurious Features.} Models suffering from SC learn spurious features far faster than core features. Consequently, early in training, models rely almost \textit{entirely on the spurious features for their predictions}. Group inference (GI) methods exploit this observation in various ways to infer groups. However, when the spurious feature cannot be learned as quickly, the model partially relies on core features for its predictions, even early in training. As a result, the underlying assumption behind existing GI methods breaks, and they can no longer infer groups accurately. We confirm this in Fig. \ref{fig:group_infer_slow} by showing the fraction of minority group examples that GI methods misclassify as belonging to majority groups.

\begin{table*}[!h]
  \centering
  \caption{{Statistics of the six datasets used in our benchmark. 
  }}\label{table:stats}
  \renewcommand{\arraystretch}{0.9}
  \scriptsize 
  \begin{tabularx}{\textwidth}{X *{6}{r}| *{6}{r}}
    \toprule
    Dataset & \multicolumn{2}{c}{Waterbirds} & \multicolumn{2}{c}{CelebA} & \multicolumn{2}{c|}{UrbanCars} & \multicolumn{2}{c}{SpuCoAnimals} & \multicolumn{2}{c}{SpuCoSun (Hard)} & \multicolumn{2}{c}{SpuCoSun (Easy)} \\
    \midrule
    \# Classes & \multicolumn{2}{c}{2} & \multicolumn{2}{c}{2}& \multicolumn{2}{c|}{2}& \multicolumn{2}{c}{4}& \multicolumn{2}{c}{4}& \multicolumn{2}{c}{4}\\
    \# Groups & \multicolumn{2}{c}{4} & \multicolumn{2}{c}{4}& \multicolumn{2}{c|}{8}& \multicolumn{2}{c}{8}& \multicolumn{2}{c}{16}& \multicolumn{2}{c}{16}\\
    \midrule
    Train & \multicolumn{2}{c}{4795} & \multicolumn{2}{c}{162770}& \multicolumn{2}{c|}{7989}& \multicolumn{2}{c}{42000}& \multicolumn{2}{c}{28092}& \multicolumn{2}{c}{28092}\\
    Val. & \multicolumn{2}{c}{1199} & \multicolumn{2}{c}{19867}& \multicolumn{2}{c|}{999}& \multicolumn{2}{c}{2100}& \multicolumn{2}{c}{1612}& \multicolumn{2}{c}{1612}\\
    Test & \multicolumn{2}{c}{5794} & \multicolumn{2}{c}{19962}& \multicolumn{2}{c|}{1000}& \multicolumn{2}{c}{4000}& \multicolumn{2}{c}{4012}& \multicolumn{2}{c}{4012}\\
    \midrule
    Class Ratio & \multicolumn{2}{c}{76.8:23.2} & \multicolumn{2}{c}{85:15} & \multicolumn{2}{c|}{50:50} & \multicolumn{2}{c}{25:25:25:25} & \multicolumn{2}{c}{25:25:25:25}& \multicolumn{2}{c}{25:25:25:25} \\
    Train Maj.:Min. & \multicolumn{2}{c}{95:5} & 52:48 & 94:6& \multicolumn{2}{c|}{90.25:4.75:0.25}& \multicolumn{2}{c}{95:5}& \multicolumn{2}{c}{98.5:0.5:0.5:0.5}& \multicolumn{2}{c}{98.5:0.5:0.5:0.5}\\
    Test Maj.:Min. & \multicolumn{2}{c}{95:5} & 52:48 & 94:6& \multicolumn{2}{c|}{25:25:25}& \multicolumn{2}{c}{50:50}& \multicolumn{2}{c}{50:50}& \multicolumn{2}{c}{50:50}\\
    \bottomrule
  \end{tabularx}
\end{table*}

\section{Higher Number of Groups \& Classes} \label{sec:groups_classes}

In this section, we study the effect of increasing the number of groups and classes on the effectiveness of existing methods in mitigating spurious correlations. We first present existing datasets and then introduce a new dataset with more classes and groups. It is important to note that conducting a controlled experiment in this context is fundamentally impossible, as altering the number of classes and groups necessarily changes other properties of the data (e.g., the learning speed of spurious or true class features). Nonetheless, our benchmarking of existing methods, across datasets with varying numbers of groups and classes, reveals trends that underscore the unsolved challenges posed by datasets with more groups and/or classes.


\subsection{Datasets \& Models}\label{sec:groups_classes:data}

We consider standard benchmark datasets for spurious correlations, namely Waterbirds and CelebA, both consisting of 2 classes and 4 groups. We also include the recently introduced UrbanCars dataset, which contains 2 classes and 8 groups. Additionally, we consider \sun (described in Sec. \ref{sec:spu_diff:data}), which has 4 classes and 16 groups. Furthermore, we introduce \animals, a new dataset with 4 classes and 8 groups.

\textbf{SpuCoAnimals (8 groups, 4 classes).} \textsc{SpuCoAnimals} is a large-scale vision dataset curated from ImageNet \cite{imagenet_cvpr09} with 4 classes: \{\texttt{landbirds}, \texttt{waterbirds}, \texttt{small dogs}, \texttt{big dogs}\}, 
\texttt{Waterbirds} and \texttt{landbirds} are spuriously correlated with \textit{water} (lake, river, sea) and, \textit{land} (grass, forest and tree) backgrounds, respectively. \texttt{Small dogs} and \texttt{big dogs} are spuriously correlated with \textit{indoor} (bed, couch and floor) and \textit{outdoor} (grass, park and road) backgrounds, respectively. 
The training set has 10500 examples per class with a 20:1 ratio between majority and minority group for each class. The validation set has 525 examples for each class with balanced groups.
The test set has 500 examples per group. 
Fig. \ref{fig:spucoanimals_data} shows examples of images from \animals. 

\textbf{Model: CLIP-pretrained ResNet50.} 
Prior work \cite{liu2021just,creager2021environment,yang2023identifying,kirichenko2023last,sagawa2020distributionally} use an ImageNet pretrained ResNet50 \cite{he2016deep} for group inference and robust training. 
To maintain consistency, we use the same setting for \water, \celeba and \cars. But, for all our datasets, we use a pretrained CLIP-ResNet50 to avoids potential data leakage from the ImageNet training set, as ImageNet-pretrained models may have already been trained on some of the images in our datasets. 

\subsection{Results and Analysis}

Table \ref{tab:all_tab} shows the Worst-Group (WG) and Average (AVG) accuracy of all methods, on the above datasets.  

\textbf{Group-Balancing outperforms Group-DRO} Interestingly, we find that across all datasets, simple Group-Balancing (GB) can achieve higher WG accuracy than the more sophisticated Group-DRO method. Group-DRO not only samples group-balanced batches during training but also computes a weighted loss over groups, assigning higher weights to worse-performing groups. The improvement of GB over Group-DRO is even more pronounced on datasets with more than 2 groups and/or more than 2 classes. This is likely due to the optimization challenges introduced by the frequent changes in the loss weights for different groups. \looseness=-1

\begin{table*}[!t]
\caption{Effect of Number of Groups / Classes: WG \& AVG accuracy ($\%$) of training with SOTA algorithms. With more groups, accuracy disparity increases. With more classes, average accuracy is harder to maintain \& there's a large spread in performance across methods}
\label{tab:all_tab}
\vspace{-3mm}
\begin{center}
\resizebox{1\textwidth}{!}{%
\begin{tabular}{|l|cc|cc|cc|cc|cc|}
\hline
 & \multicolumn{2}{c|}{\cellcolor{cyan!10}\textsc{Waterbirds}} & \multicolumn{2}{c|}{\cellcolor{cyan!10}\textsc{CelebA}} & \multicolumn{2}{c|}{\cellcolor{gray!10}\textsc{UrbanCars}} & \multicolumn{2}{c|}{\cellcolor{orange!10}\textsc{SpuCoAnimals}} & \multicolumn{2}{c|}{\cellcolor{red!10}\textsc{SpuCoSun}} \\
 & \cellcolor{cyan!10}WG & \cellcolor{cyan!10}AVG  & \cellcolor{cyan!10}WG & \cellcolor{cyan!10}AVG & \cellcolor{gray!10}WG & \cellcolor{gray!10}AVG & \cellcolor{orange!10}WG & \cellcolor{orange!10}AVG & \cellcolor{red!10}WG & \cellcolor{red!10}AVG \\\hline\hline
 ERM & \cellcolor{cyan!10}$80.1_{\pm0.7}$ & \cellcolor{cyan!10}$97.8_{\pm 0.3}$ & \cellcolor{cyan!10}$78.7_{\pm 2.1}$ & \cellcolor{cyan!10}$83.4_{\pm 3.5}$ & \cellcolor{gray!10}$56.8_{\pm 5.6}$ & \cellcolor{gray!10}$96.4_{\pm 0.5}$ & \cellcolor{orange!10}$39.0_{\pm 1.4}$ & \cellcolor{orange!10}$75.8_{\pm 1.8}$ & \cellcolor{red!10}$29.3_{\pm 1.4}$& \cellcolor{red!10}$96.0_{\pm 0.2}$\\\hline\hline
 JTT & \cellcolor{cyan!10}$83.1_{\pm 3.5}$ & \cellcolor{cyan!10}$90.6_{\pm 0.3}$ & \cellcolor{cyan!10}$81.5_{\pm 1.7}$ & \cellcolor{cyan!10}$88.1_{\pm 0.3}$ & \cellcolor{gray!10}$68.3_{\pm 6.7}$ & \cellcolor{gray!10}$91.9_{\pm 4.4}$ & \cellcolor{orange!10}$57.4_{\pm 2.3}$ & \cellcolor{orange!10}$83.6_{\pm 1.1}$ & \cellcolor{red!10}$55.8_{\pm 1.1}$ & \cellcolor{red!10}$88.8_{\pm 0.6}$\\
 \spare & \cellcolor{cyan!10}$91.6_{\pm 0.8}$ & \cellcolor{cyan!10}$96.2_{\pm 0.6}$ & \cellcolor{cyan!10}$86.5_{\pm3.3}$ & \cellcolor{cyan!10}$89.8_{\pm0.3}$ & \cellcolor{gray!10}$80.5_{\pm 3.9}$ & \cellcolor{gray!10}$90.3_{\pm 2.5}$ & \cellcolor{orange!10}$61.6_{\pm 2.4}$ & \cellcolor{orange!10}$76.5_{\pm 1.1}$ & \cellcolor{red!10}$62.2_{\pm 1.4}$ & \cellcolor{red!10}$80.3_{\pm 1.8}$\\
 EIIL & \cellcolor{cyan!10}$83.5_{\pm 2.8}$ & \cellcolor{cyan!10}$94.2_{\pm 1.3}$ & \cellcolor{cyan!10}$78.9_{\pm 0.6}$ & \cellcolor{cyan!10}$86.9_{\pm 2.0}$ & \cellcolor{gray!10}$78.9_{\pm 2.8}$ & \cellcolor{gray!10}$90.2_{\pm 1.3}$ & \cellcolor{orange!10}$64.9_{\pm 3.5}$ & \cellcolor{orange!10}$77.3_{\pm 2.4}$ & \cellcolor{red!10}$65.9_{\pm 1.3}$& \cellcolor{red!10}$76.1_{\pm 3.1}$\\
 \hline\hline
 SSA & \cellcolor{cyan!10}$85.1_{\pm 1.3}$ & \cellcolor{cyan!10}$96.7_{\pm 1.6}$ & \cellcolor{cyan!10}$89.4_{\pm 0.5}$ & \cellcolor{cyan!10}$91.2_{\pm 0.2}$ & \cellcolor{gray!10}$78.7_{\pm 2.6}$ & \cellcolor{gray!10}$93.4_{\pm 0.7}$ & \cellcolor{orange!10}$64.4_{\pm 3.1}$ & \cellcolor{orange!10}$75.5_{\pm 1.4}$ & \cellcolor{red!10}$68.7_{\pm 0.8}$ & \cellcolor{red!10}$83.9_{\pm 0.8}$\\ 
 $\text{DFR}^{\text{Val}}_{\text{Tr}}$ & \cellcolor{cyan!10}$90.6_{\pm 0.4}$ & \cellcolor{cyan!10}$93.4_{\pm 0.2}$ & \cellcolor{cyan!10}$89.1_{\pm 1.2}$ & \cellcolor{cyan!10}$91.3_{\pm 0.1}$ & \cellcolor{gray!10}$81.9_{\pm 1.2}$ & \cellcolor{gray!10}$92.5_{\pm 1.8}$ & \cellcolor{orange!10}$66.5_{\pm 1.8}$ & \cellcolor{orange!10}$80.0_{\pm 3.7}$ & \cellcolor{red!10}$67.3_{\pm 1.6}$ & \cellcolor{red!10}$79.9_{\pm 2.6}$\\
 \hline\hline
 GB & \cellcolor{cyan!10}$88.1_{\pm 0.6}$ & \cellcolor{cyan!10}$95.5_{\pm 0.4}$ & \cellcolor{cyan!10}$88.6_{\pm 1.0}$ & \cellcolor{cyan!10}$91.1_{\pm 0.1}$ & \cellcolor{gray!10}$82.7_{\pm 2.0}$ & \cellcolor{gray!10}$92.8_{\pm 0.0}$ & \cellcolor{orange!10}$69.9_{\pm 0.1}$ & \cellcolor{orange!10}$79.1_{\pm 0.1}$ & \cellcolor{red!10}$69.9_{\pm 0.6}$ & \cellcolor{red!10}$79.1_{\pm 0.2}$\\
 GDRO & \cellcolor{cyan!10}$85.7_{\pm 0.8}$ & \cellcolor{cyan!10}$94.3_{\pm 1.0}$ & \cellcolor{cyan!10}$89.4_{\pm 0.1}$ & \cellcolor{cyan!10}$91.4_{\pm 0.1}$ & \cellcolor{gray!10}$78.9_{\pm 2.6}$ & \cellcolor{gray!10}$92.7_{\pm 1.6}$ & \cellcolor{orange!10}$67.1_{\pm 0.2}$ & \cellcolor{orange!10}$75.1_{\pm 0.1}$ & \cellcolor{red!10}$65.5_{\pm 2.3}$ & \cellcolor{red!10}$78.2_{\pm 0.8}$\\
 PDE & \cellcolor{cyan!10}$90.3_{\pm{0.3}}$ & \cellcolor{cyan!10}$92.4_{\pm{0.8}}$ & \cellcolor{cyan!10}$91.0_{\pm{0.4}}$ & \cellcolor{cyan!10}$92.0_{\pm{0.6}}$ & \cellcolor{gray!10}$84.3_{\pm 1.7}$ & \cellcolor{gray!10}$92.2_{\pm 2.0}$ & \cellcolor{orange!10}$70.1_{\pm 2.0}$ & \cellcolor{orange!10}$78.7_{\pm 1.0}$ & \cellcolor{red!10}$67.6_{\pm 0.8}$ & \cellcolor{red!10}$77.7_{\pm 1.2}$\\ \hline
 \multicolumn{1}{c}{} & \multicolumn{4}{c}{\textcolor{cyan}{\textbf{4 groups, 2 classes}}} & \multicolumn{2}{c}{\textcolor{gray}{\textbf{8 groups, 2 classes}}} & \multicolumn{2}{c}{\textcolor{orange}{\textbf{8 groups, 4 classes}}} & \multicolumn{2}{c}{\textcolor{red}{\textbf{16 groups, 4 classes}}}\\
\end{tabular}%
}
\end{center}
\end{table*}


\begin{figure}[h]
    \centering
    \includegraphics[width=0.6\linewidth]{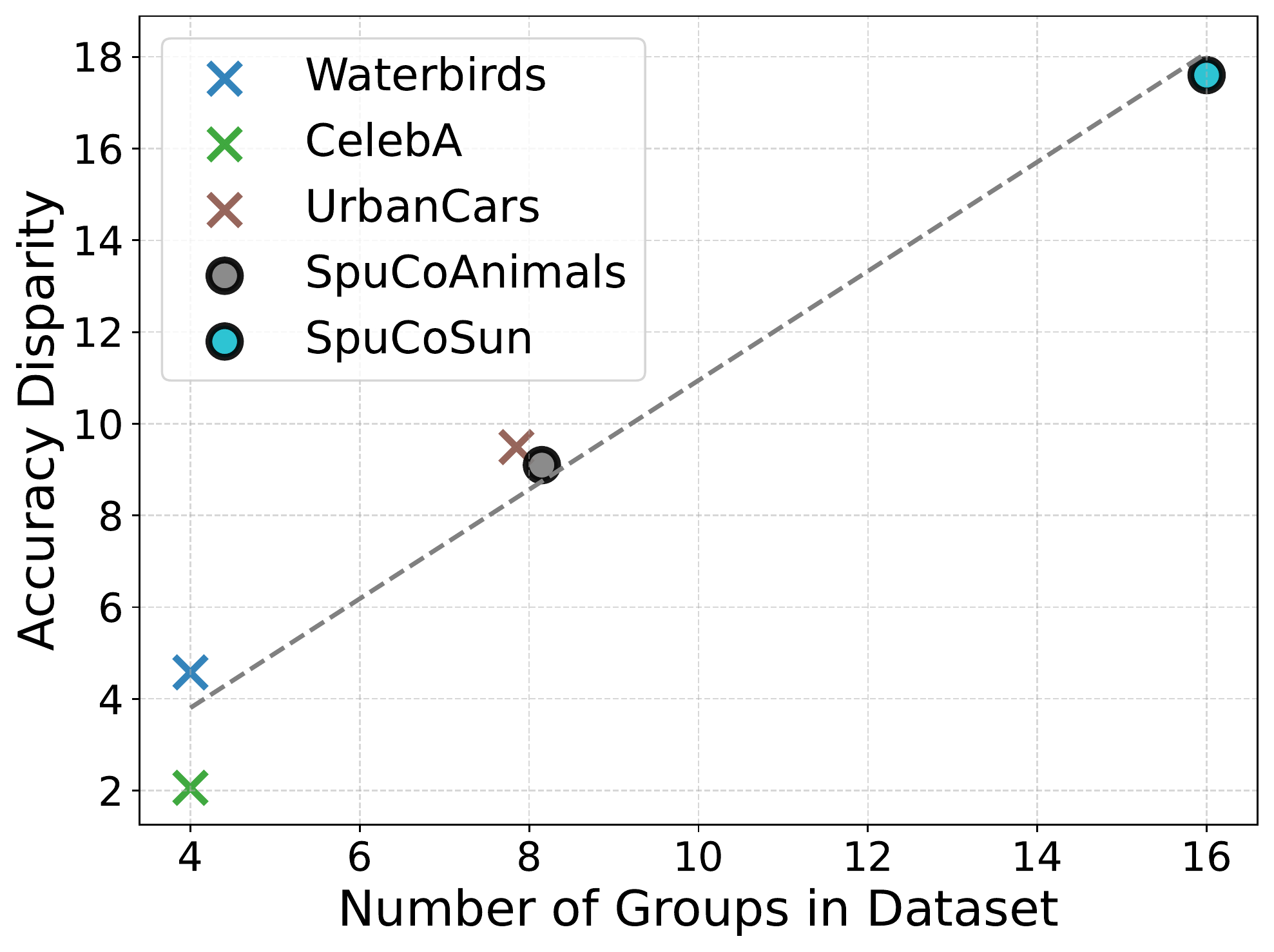} 
    \caption{Accuracy disparity increases with \# of groups}
    \label{fig:gap_vs_num_groups}
     \includegraphics[width=0.7\linewidth]{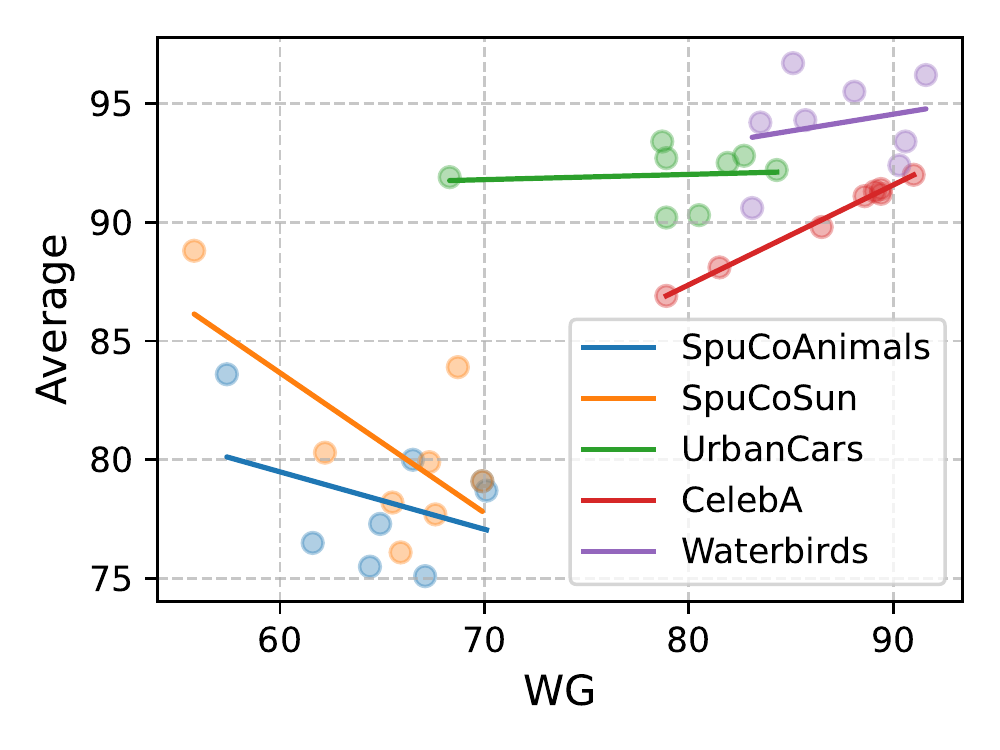} 
     \vspace{-3mm}
    \caption{AVG vs. WG accuracy, averaged over different methods on each dataset. Waterbirds, CelebA, UrbanCars, \sun and \animal have 2, 2, 2, 4, 4 classes, respectively. As \# classes increases more classes, improving WG accuracy more negatively affects AVG accuracy and the peformance spread across methods widens. 
    } 
    \label{fig:var_avg_error}
\end{figure}

\vspace{-4mm}
\paragraph{Accuracy Disparity is Larger with More Groups}
\begin{figure}[h]
    \includegraphics[width=1.0\linewidth]{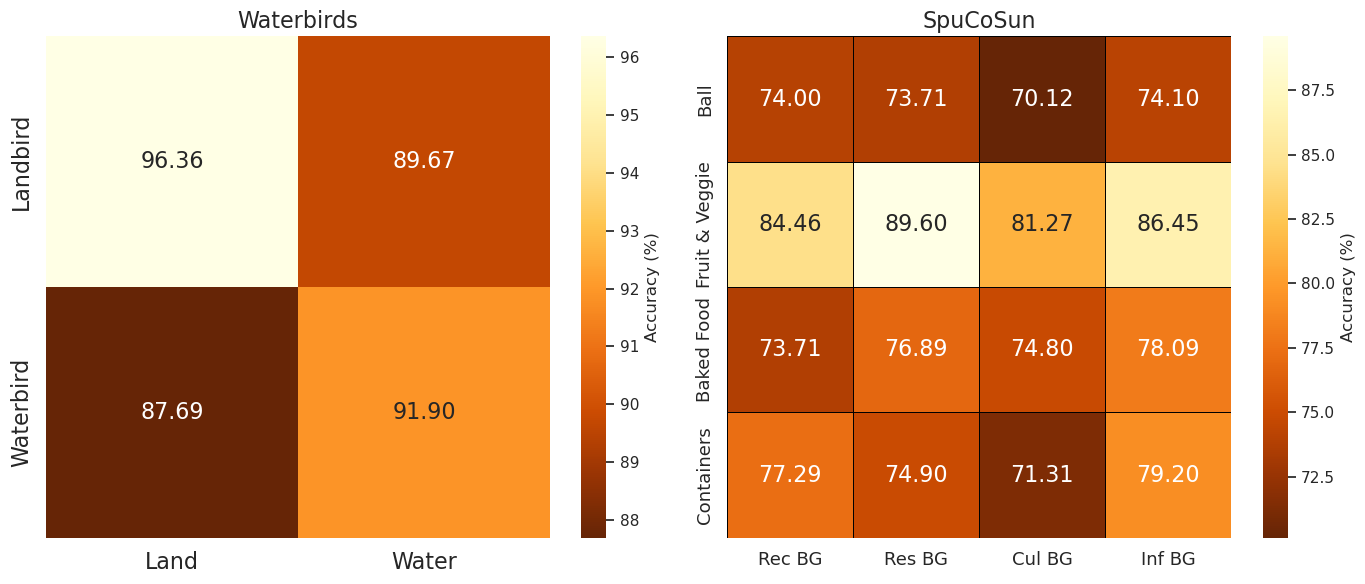} 
    \caption{Heatmap of Group accuracies for Waterbirds and \sun (Darker colors represent groups with lower accuracies)}
    \label{fig:group_acc}
    \includegraphics[width=\linewidth]{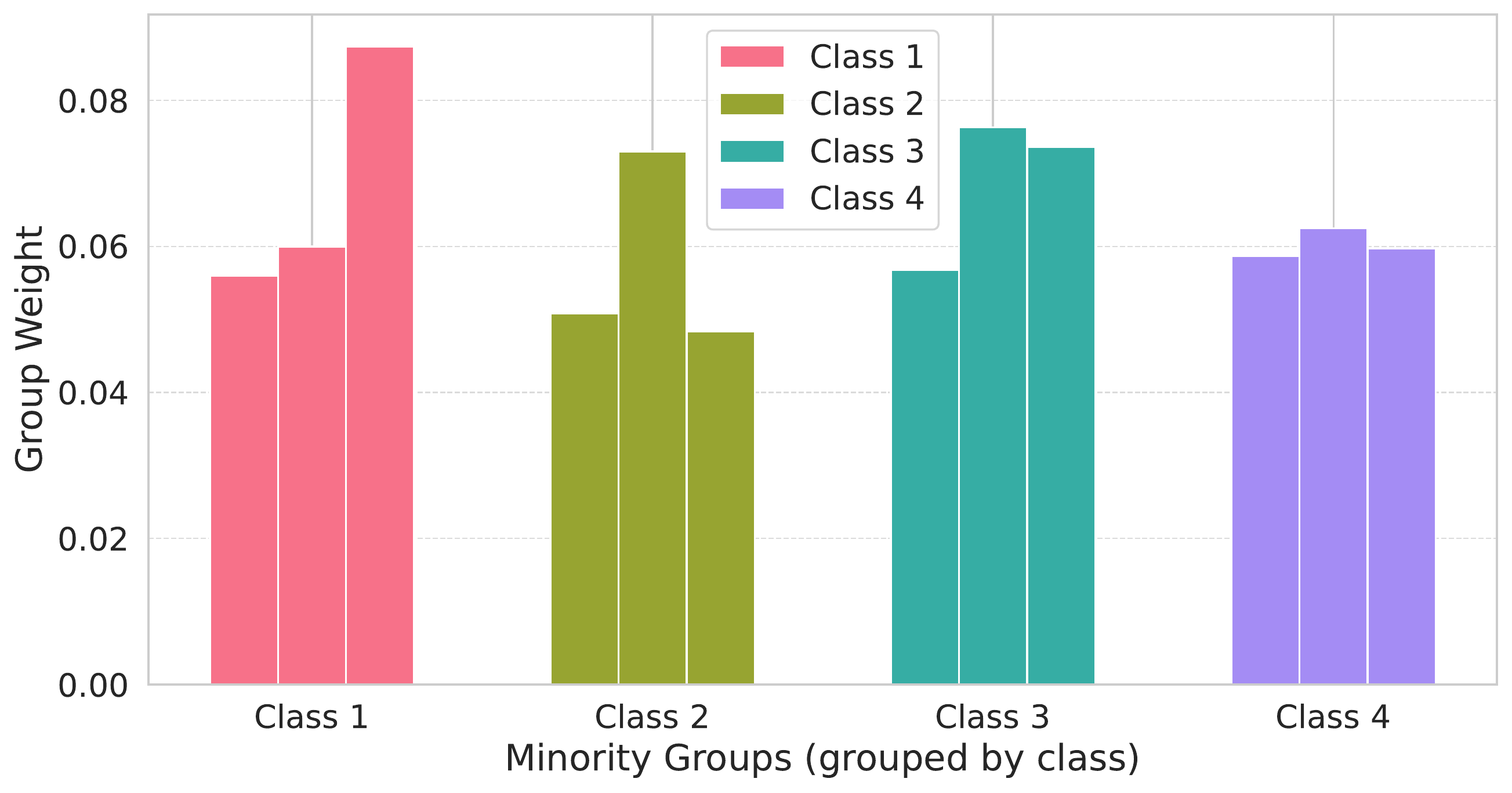}
    \caption{Group Weights for Minority Groups Computed by GroupDRO on \sun (end of training)}\label{fig:group_dro}
\end{figure}
Figure \ref{fig:gap_vs_num_groups} shows the accuracy disparity, i.e., the difference between WG and AVG accuracy, averaged across all methods for datasets with varying numbers of groups. We observe that the accuracy disparity increases almost linearly with the number of groups. This confirms that it becomes significantly more challenging for existing methods to reduce the accuracy disparity as the number of groups increases. Specifically, for \water and \celeba, datasets with only 4 groups, the average accuracy disparity is relatively small, at no more than 5\%. When the number of groups doubles to 8 in \cars and \animals, the accuracy disparity also doubles, reaching nearly 10\%. Finally, on \sun, which has 16 groups, the accuracy disparity is as high as 18\%. Delving deeper into the results in Table \ref{tab:all_tab}, we see that even the best-performing method on each dataset struggles considerably as the number of groups increases. For example, the accuracy disparity for the best method, PDE, is less than 2.1\% on datasets with 4 groups. However, as the number of groups increases from \cars and \animals to \sun (Hard), the accuracy disparity for the best method rises to 7.9\%, 8.6\%, and 10.1\%, respectively. This increase in accuracy disparity is primarily caused by the difficulty of model selection—specifically, selecting optimal hyperparameters based on validation set WG accuracy—for datasets with more groups. As shown in Fig. \ref{fig:group_acc}, where we visualize the group-wise accuracy of simple Group Balancing on \sun and \water, datasets with more groups have a larger number of minority groups with lower accuracies. This makes it challenging to find a set of hyperparameters that optimizes performance across all poorly performing groups. Moreover, for robust training methods like GDRO, which weight losses for different groups based on their performance, the larger number of groups further complicates optimization. Fig. \ref{fig:group_dro} shows that the loss weights for different minority groups, even within the same class, vary significantly. This occurs despite all minority groups within a class having an equal number of samples. Notably, such imbalanced weighting among minority groups is not possible in datasets like \water and \celeba, which have only a single minority group per class. This imbalance likely leads to overfitting to one of the minority groups rather than learning class features in a more generalizable manner. Addressing the larger accuracy disparity for datasets with more groups remains an unsolved challenge for methods tackling spurious correlations.

\paragraph{Maintaining a High Avg Acc when Data Has More Classes Is Difficult.}
Fig. \ref{fig:var_avg_error} illustrates an interesting observation: when the data contains more classes, improving the worst-group (WG) accuracy negatively impacts the average (AVG) accuracy. Even when the number of groups is controlled (both \cars and \animals have 8 groups), increasing the number of classes (from 2 in \cars to 4 in \animals) shifts the correlation between WG and AVG accuracy from near zero to strongly negative, with a large spread in AVG accuracy across methods. Most state-of-the-art (SOTA) methods address spurious correlations (after group inference, if necessary) by upsampling identified minority group examples. The model then relies heavily on these upsampled examples to learn the true class feature. With more classes, as the classification problem becomes more challenging, it becomes increasingly critical to upsample examples with representative true class features, rather than outliers. The strong negative correlation and the large spread in AVG accuracy across methods (see Fig. \ref{fig:var_avg_error}) indicate that it is no longer sufficient to merely identify and upsample minority examples. Instead, it is essential to upsample minority examples that enable the model to generalizably learn the true class feature. This underscores a new challenge: maintaining high AVG accuracy while improving WG accuracy in the presence of more classes.

\section{Model Selection}\label{sec:model_sel}

Model selection criteria determine which hyperparameters to use to obtain the best worst-group accuracy at test time. \citet{domainbed} has studied and shown the importance of model selection criteria for domain generalization methods. We first make a similar observation for methods for spurious correlations and highlight how this is especially true with more groups and/or classes. Then, we demonstrate how effective model selection can help significantly when addressing multiple spurious features. After we have established the importance of model selection, we explore potential alternatives for reducing the high costs associated with model selection. 

\begin{figure}
    \centering
    \includegraphics[width=\linewidth]{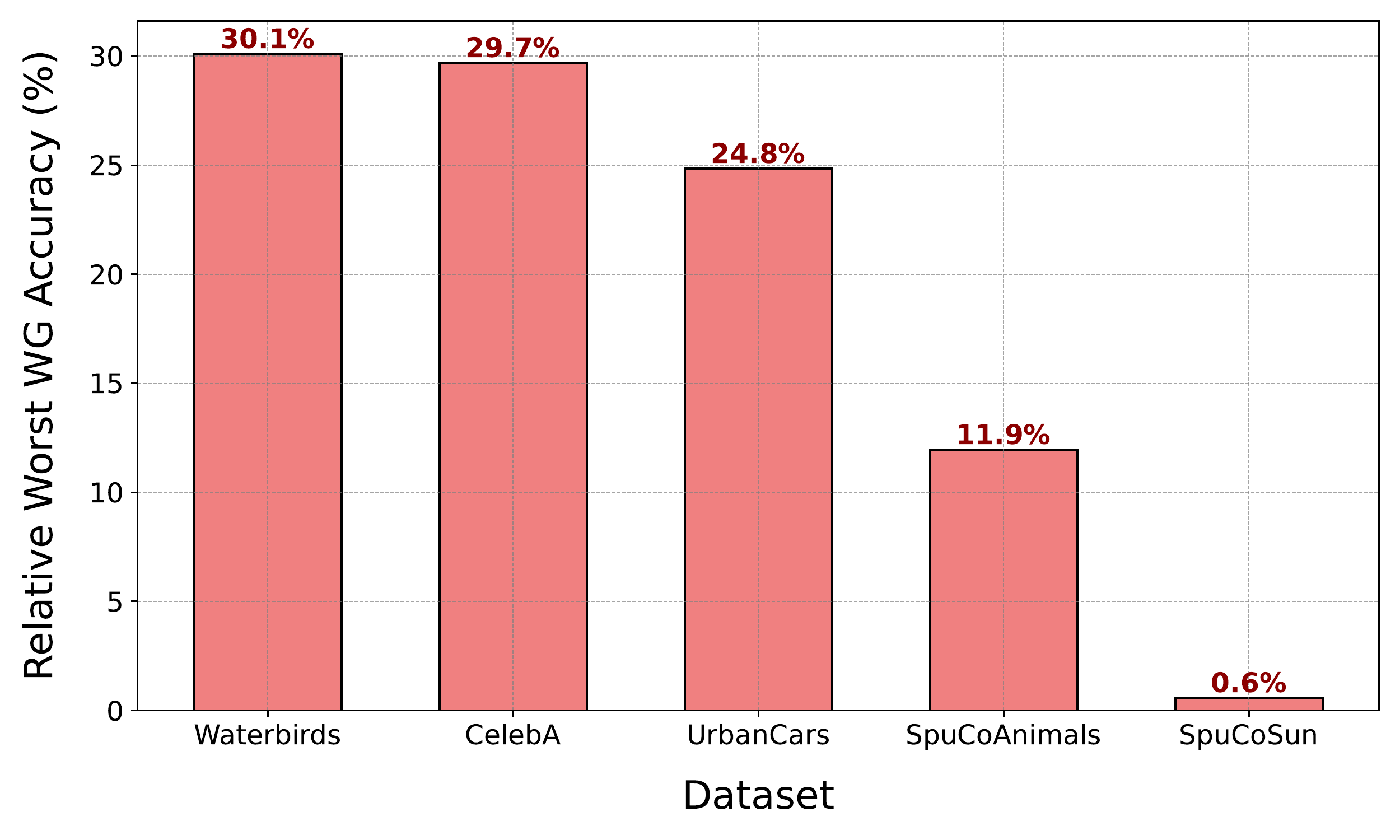}
    \caption{Normalized worst hyperparameter WG accuracy relative to best hyperparameter WG accuracy, averaged across methods and datasets. The low relative performance of the worst hyperparameters highlights the model's high sensitivity to hyperparameter selection.}
    \label{fig:tuning_single}
\end{figure}
\vspace{-4mm}

\paragraph{All Methods Are Highly Sensitive to Model Selection} Fig. \ref{fig:tuning_single} shows the WG accuracy of the worst hyperparameters, normalized by the WG accuracy of the best hyperparameters, across different datasets for all SOTA algorithms. For each method on existing datasets, we tuned based on the recommended hyperparameter ranges; for our datasets, we list our tuning ranges in Sec. \ref{sec:hyper_range}.
We observe that across all datasets, the worst hyperparameter combinations within the recommended ranges achieve at most 30\% of the WG accuracy of the best hyperparameters. The performance gap due to poor model selection is even more pronounced on datasets with more groups and/or classes (\cars, \animals, \sun). Thus, a suboptimal choice of hyperparameters can significantly reduce method performance, necessitating thorough and costly model selection processes, which undermine the practicality of existing methods. \looseness=-1

\begin{figure}
    \centering
    \includegraphics[width=\linewidth]{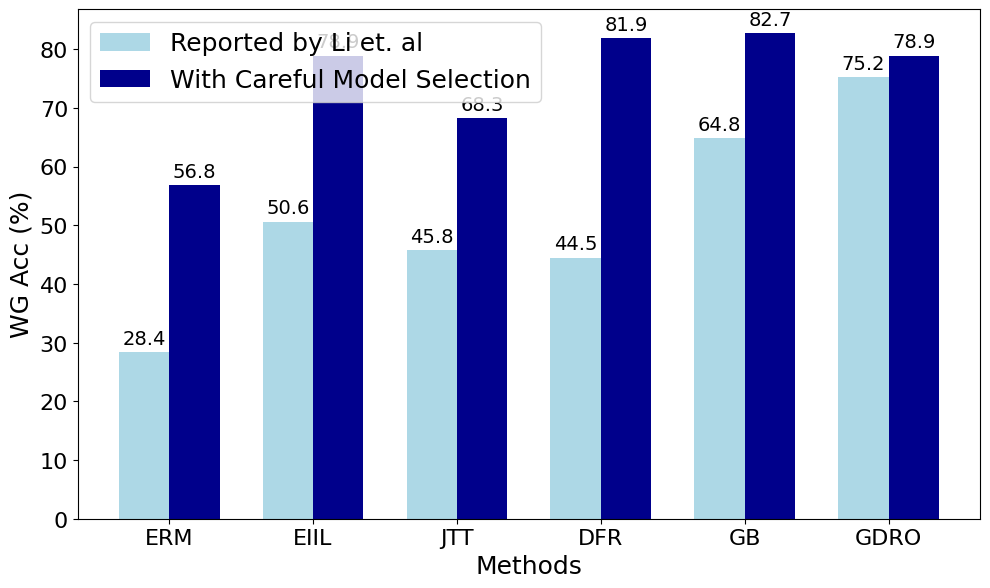}
    \caption{Showing Large Improvements on \cars, Achieved through Careful Model Selection}
    \label{fig:model_sel:cars_improve}
\end{figure}

\paragraph{Model Selection in the Presence of Multiple Spurious Features}
\citet{Li_2023_CVPR_Whac_A_Mole} proposed \cars, a dataset containing multiple spurious features, and highlighted the challenge of improving WG accuracy in this setting. They demonstrated that improving the performance of minority groups with respect to one spurious feature often leads to a decrease in performance for minority groups corresponding to another spurious feature. Consequently, \citet{Li_2023_CVPR_Whac_A_Mole} observed consistently poor WG accuracy on \cars across various methods. Here, we show that careful model selection can significantly mitigate this issue. For each method, we perform model selection by randomly sampling 16 hyperparameter combinations from the ranges specified for the respective method. Figure~\ref{fig:model_sel:cars_improve} compares the performance of SOTA algorithms as reported by \citet{Li_2023_CVPR_Whac_A_Mole} on \cars with the results obtained through our model selection. Careful model selection achieves an average improvement of 23\%, with gains as high as 37\%.

\begin{figure}
    \centering
    \includegraphics[width=1\linewidth]{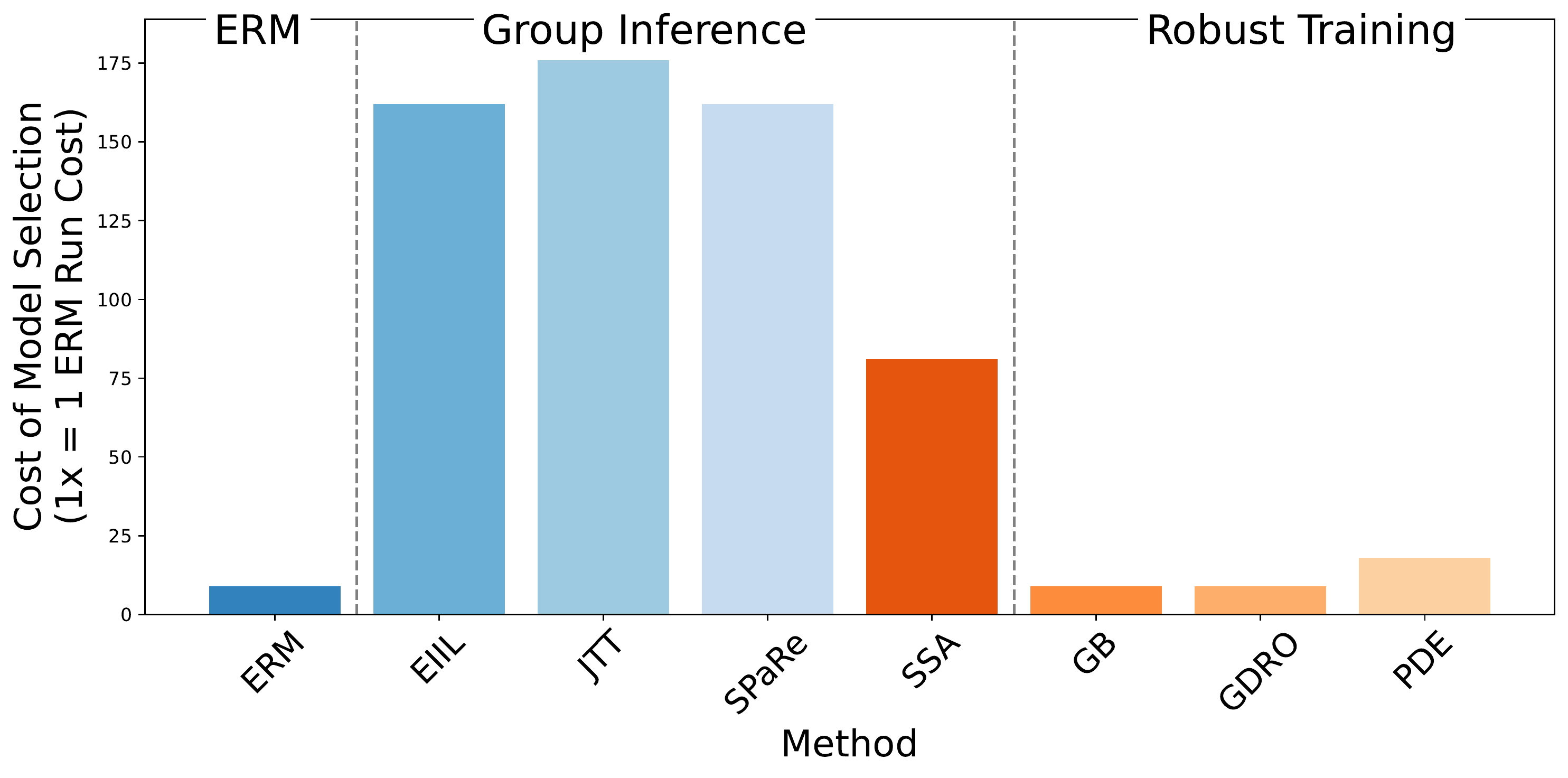}
    \caption{Highlighting the High Cost of Model Selection for Group Inference Methods}
    \label{fig:model_sel_cost}
\end{figure}

\begin{figure}
    \centering
    \begin{subfigure}[b]{0.3\linewidth}
        \centering
        \includegraphics[width=\linewidth]{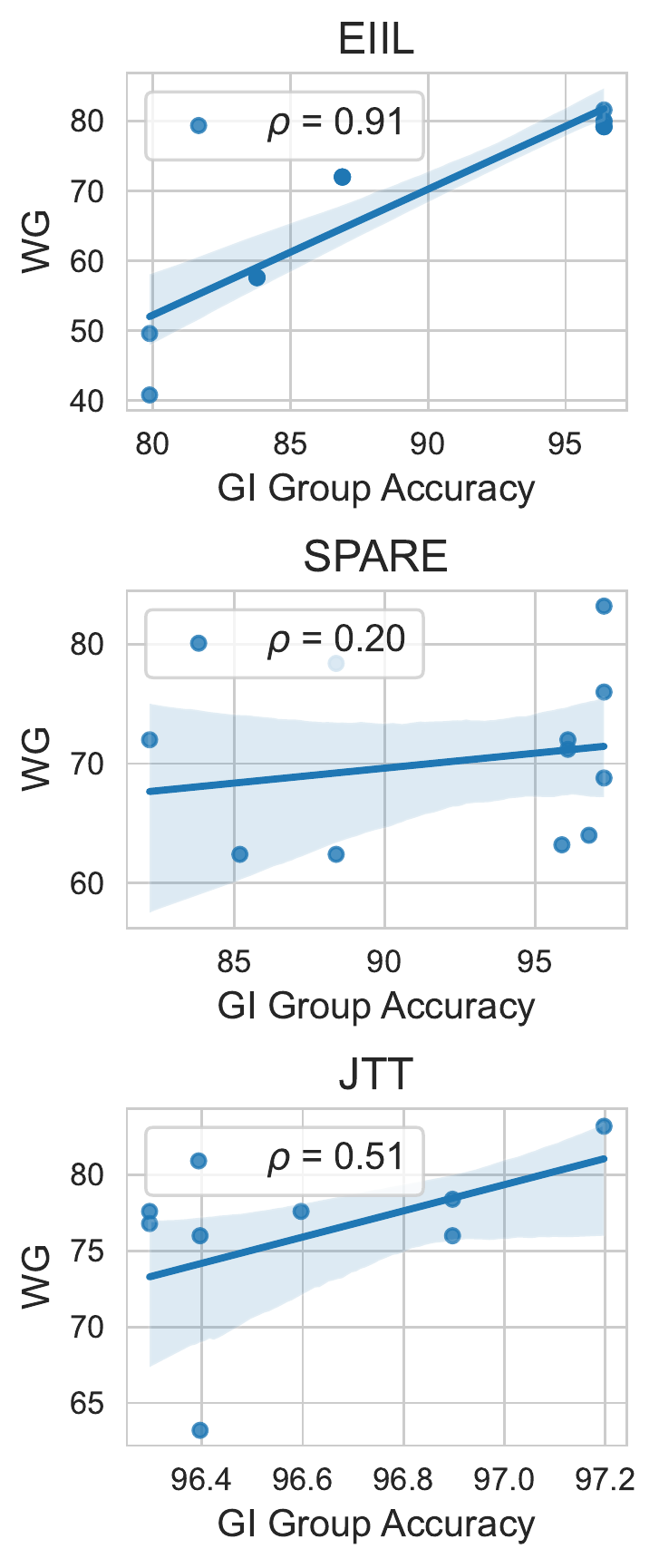}
        \caption{\cars}
        \label{fig:model_sel:gi:cars}
    \end{subfigure}%
    \hfill
    \begin{subfigure}[b]{0.3\linewidth}
        \centering
        \includegraphics[width=\linewidth]{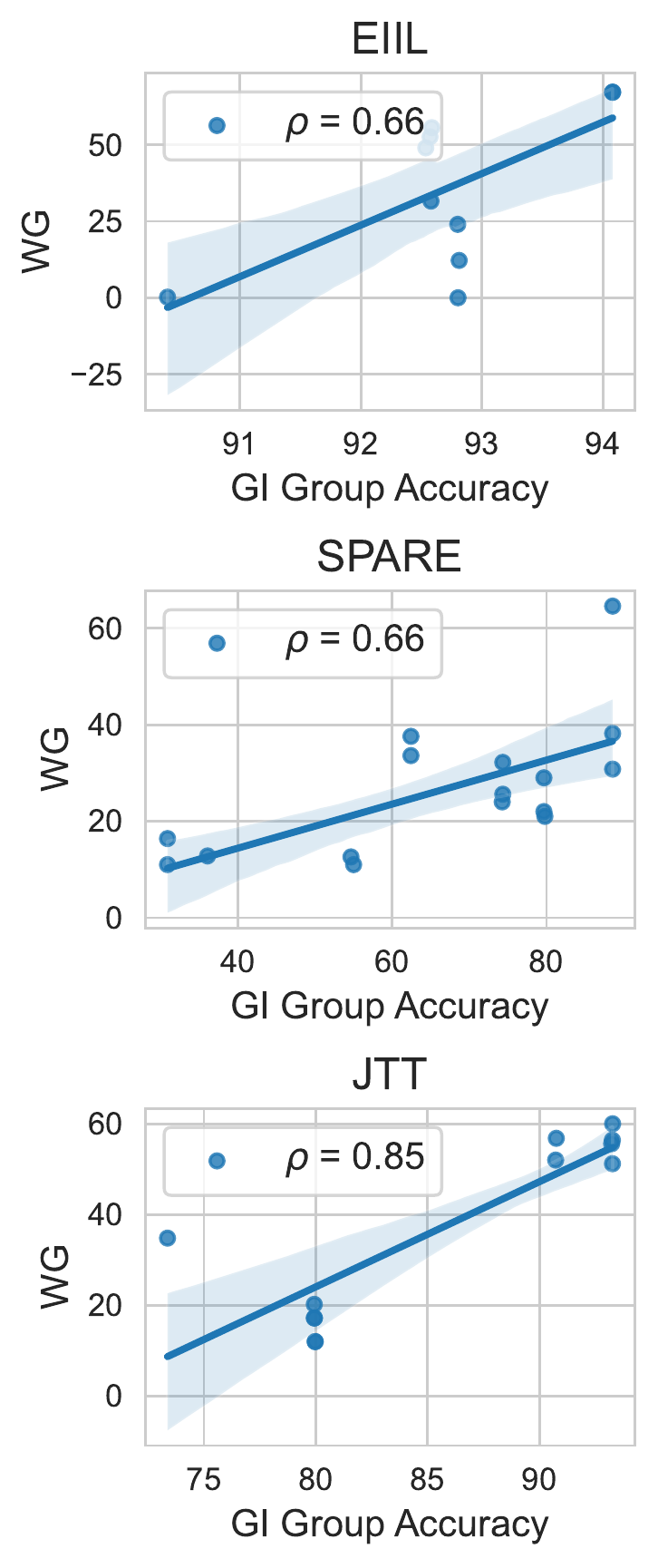}
        \caption{\animals}
        \label{fig:model_sel:gi:animals}
    \end{subfigure}%
    \hfill
    \begin{subfigure}[b]{0.3\linewidth}
        \centering
        \includegraphics[width=\linewidth]{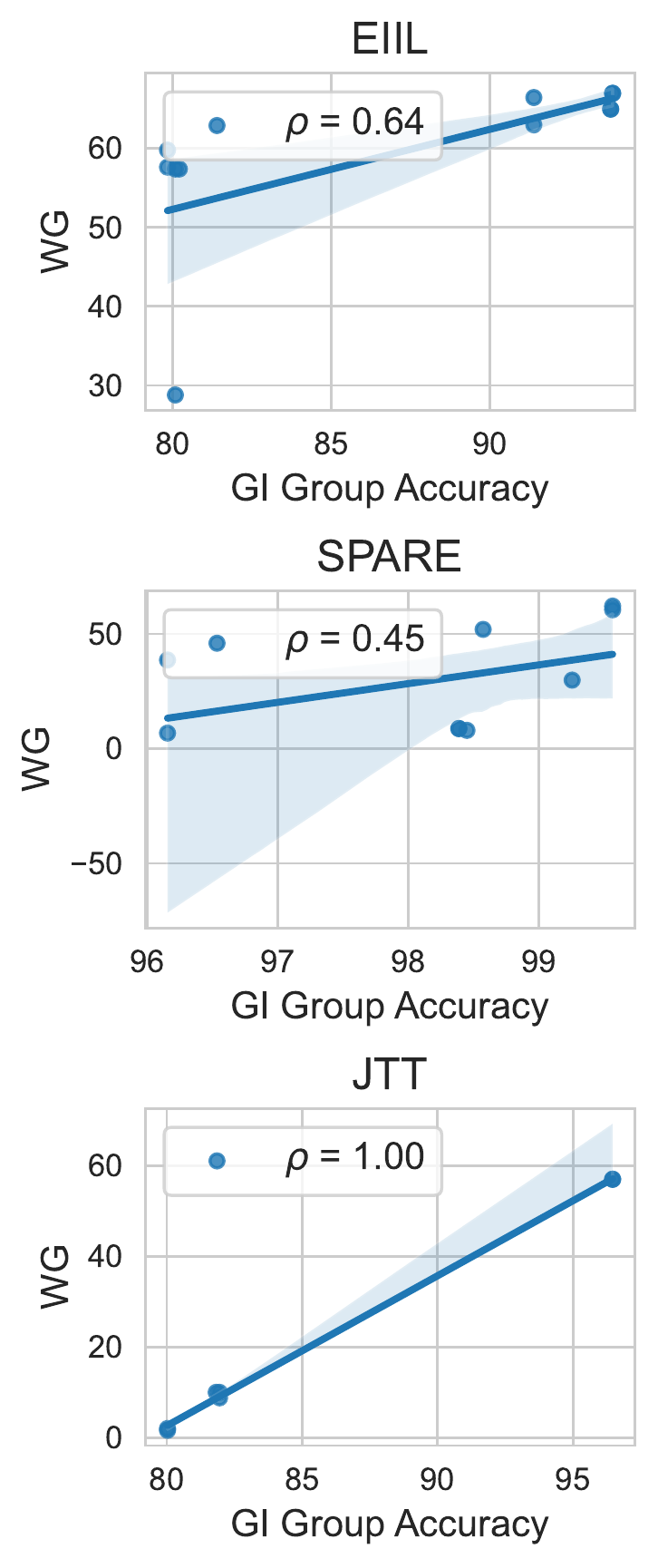}
        \caption{\sun}
        \label{fig:model_sel:gi:sun}
    \end{subfigure}
    \caption{Positive Correlations between Accuracy of Inferred Groups and Resulting WG Accuracy}
    \label{fig:model_sel:gi}
\end{figure}

\vspace{-4mm}
\paragraph{Towards Efficient Model Selection for Group Inference Methods} 
Fig. \ref{fig:model_sel_cost} shows that model selection is 100x more expensive for group inference methods, compared to others. This is due to the fact that for every potential combination of hyperparameters, these methods must infer groups, then train a model using the inferred groups, and then use the resulting validation worst-group accuracy as the model selection criterion. In many real-world scenarios, we do rely on these methods since group labels are not available for training data and must be inferred.
To remedy the very high costs of model selection for group inference methods, we investigate directly evaluating the quality of the inferred groups on the group-labeled validation set, by framing group inference as \textit{majority} or \textit{minority} as a classification problem (henceforth, referred to as \textit{group inference accuracy}). Fig. \ref{fig:model_sel:gi} shows strong positive correlations of \textit{group inference accuracy}, across group inference methods, with the final test worst-group accuracy. While these correlations are slightly weaker for SPARE, the trend remains consistent. Thus, we find that directly evaluating inferred groups is a promising approach for far more cost-efficient model selection for group inference methods. 

\section{Conclusion}

We introduced three previously unexplored settings: 1) a larger number of classes, 2) a greater number of groups, and 3) slower-learned spurious features that pose challenges for existing methods aimed at improving worst-group (WG) accuracy in the presence of spurious features. Through extensive experiments, spanning \textbf{8} state-of-the-art (SOTA) methods and 5 datasets, we systematically evaluate the performance of these methods in mitigating spurious correlations under these new settings, uncovering several open challenges in this area. Additionally, we demonstrate that the optimal performance reported for SOTA methods is highly sensitive to model selection and propose an efficient alternative to the costly but crucial model selection process used in group inference methods. We hope our benchmark and findings will guide both theoreticians and practitioners in developing the next generation of methods to improve worst-group accuracy in the presence of spurious correlations.

\textbf{Limitations.} While our work introduced more challenging datasets and provided a comprehensive benchmark for evaluating SOTA methods addressing spurious correlations on {vision} tasks, generalizability of our findings to other domains, such as natural language processing, remains to be studied. \looseness=-1

{
    \small
    \bibliography{references}

\begin{thebibliography}{35}
\providecommand{\natexlab}[1]{#1}
\providecommand{\url}[1]{\texttt{#1}}
\expandafter\ifx\csname urlstyle\endcsname\relax
  \providecommand{\doi}[1]{doi: #1}\else
  \providecommand{\doi}{doi: \begingroup \urlstyle{rm}\Url}\fi

\bibitem[Barocas et~al.(2023)Barocas, Hardt, and
  Narayanan]{barocas2023fairness}
Solon Barocas, Moritz Hardt, and Arvind Narayanan.
\newblock \emph{Fairness and machine learning: Limitations and opportunities}.
\newblock MIT press, 2023.

\bibitem[Chi et~al.(2021)Chi, Tian, Gordon, and Zhao]{chi2021understanding}
Jianfeng Chi, Yuan Tian, Geoffrey~J Gordon, and Han Zhao.
\newblock Understanding and mitigating accuracy disparity in regression.
\newblock In \emph{International conference on machine learning}, pages
  1866--1876. PMLR, 2021.

\bibitem[Creager et~al.(2021)Creager, Jacobsen, and
  Zemel]{creager2021environment}
Elliot Creager, J{\"o}rn-Henrik Jacobsen, and Richard Zemel.
\newblock Environment inference for invariant learning.
\newblock In \emph{International Conference on Machine Learning}, pages
  2189--2200. PMLR, 2021.

\bibitem[Deng et~al.(2009)Deng, Dong, Socher, Li, Li, and
  Fei-Fei]{imagenet_cvpr09}
J.~Deng, W.~Dong, R.~Socher, L.-J. Li, K.~Li, and L.~Fei-Fei.
\newblock {ImageNet: A Large-Scale Hierarchical Image Database}.
\newblock In \emph{CVPR09}, 2009.

\bibitem[Deng* et~al.(2023)Deng*, Yang*, Mirzasoleiman, and Gu]{deng2023robust}
Yihe Deng*, Yu~Yang*, Baharan Mirzasoleiman, and Quanquan Gu.
\newblock Robust learning with progressive data expansion against spurious
  correlation.
\newblock In A.~Oh, T.~Naumann, A.~Globerson, K.~Saenko, M.~Hardt, and
  S.~Levine, editors, \emph{Advances in Neural Information Processing Systems},
  volume~36, pages 1390--1402. Curran Associates, Inc., 2023.
\newblock URL
  \url{https://proceedings.neurips.cc/paper_files/paper/2023/file/0506ad3d1bcc8398a920db9340f27fe4-Paper-Conference.pdf}.

\bibitem[Gulrajani and Lopez-Paz(2020{\natexlab{a}})]{domainbed}
Ishaan Gulrajani and David Lopez-Paz.
\newblock In search of lost domain generalization, 2020{\natexlab{a}}.

\bibitem[Gulrajani and Lopez-Paz(2020{\natexlab{b}})]{gulrajani2020search}
Ishaan Gulrajani and David Lopez-Paz.
\newblock In search of lost domain generalization.
\newblock \emph{arXiv preprint arXiv:2007.01434}, 2020{\natexlab{b}}.

\bibitem[He et~al.(2016)He, Zhang, Ren, and Sun]{he2016deep}
Kaiming He, Xiangyu Zhang, Shaoqing Ren, and Jian Sun.
\newblock Deep residual learning for image recognition.
\newblock In \emph{Proceedings of the IEEE conference on computer vision and
  pattern recognition}, pages 770--778, 2016.

\bibitem[He et~al.(2021)He, Shen, and Cui]{he2021towards}
Yue He, Zheyan Shen, and Peng Cui.
\newblock Towards non-iid image classification: A dataset and baselines.
\newblock \emph{Pattern Recognition}, 110:\penalty0 107383, 2021.

\bibitem[Hendrycks et~al.(2021{\natexlab{a}})Hendrycks, Basart, Mu, Kadavath,
  Wang, Dorundo, Desai, Zhu, Parajuli, Guo, et~al.]{hendrycks2021many}
Dan Hendrycks, Steven Basart, Norman Mu, Saurav Kadavath, Frank Wang, Evan
  Dorundo, Rahul Desai, Tyler Zhu, Samyak Parajuli, Mike Guo, et~al.
\newblock The many faces of robustness: A critical analysis of
  out-of-distribution generalization.
\newblock In \emph{Proceedings of the IEEE/CVF international conference on
  computer vision}, pages 8340--8349, 2021{\natexlab{a}}.

\bibitem[Hendrycks et~al.(2021{\natexlab{b}})Hendrycks, Zhao, Basart,
  Steinhardt, and Song]{hendrycks2021natural}
Dan Hendrycks, Kevin Zhao, Steven Basart, Jacob Steinhardt, and Dawn Song.
\newblock Natural adversarial examples.
\newblock In \emph{Proceedings of the IEEE/CVF conference on computer vision
  and pattern recognition}, pages 15262--15271, 2021{\natexlab{b}}.

\bibitem[Idrissi et~al.(2022{\natexlab{a}})Idrissi, Arjovsky, Pezeshki, and
  Lopez-Paz]{idrissi2022simple}
Badr~Youbi Idrissi, Martin Arjovsky, Mohammad Pezeshki, and David Lopez-Paz.
\newblock Simple data balancing achieves competitive worst-group-accuracy,
  2022{\natexlab{a}}.

\bibitem[Idrissi et~al.(2022{\natexlab{b}})Idrissi, Bouchacourt, Balestriero,
  Evtimov, Hazirbas, Ballas, Vincent, Drozdzal, Lopez-Paz, and
  Ibrahim]{idrissi2022imagenet}
Badr~Youbi Idrissi, Diane Bouchacourt, Randall Balestriero, Ivan Evtimov, Caner
  Hazirbas, Nicolas Ballas, Pascal Vincent, Michal Drozdzal, David Lopez-Paz,
  and Mark Ibrahim.
\newblock Imagenet-x: Understanding model mistakes with factor of variation
  annotations.
\newblock \emph{arXiv preprint arXiv:2211.01866}, 2022{\natexlab{b}}.

\bibitem[Kim et~al.(2021)Kim, Lee, and Choo]{kim2021biaswap}
Eungyeup Kim, Jihyeon Lee, and Jaegul Choo.
\newblock Biaswap: Removing dataset bias with bias-tailored swapping
  augmentation, 2021.

\bibitem[Kirichenko et~al.(2023)Kirichenko, Izmailov, and
  Wilson]{kirichenko2023last}
Polina Kirichenko, Pavel Izmailov, and Andrew~Gordon Wilson.
\newblock Last layer re-training is sufficient for robustness to spurious
  correlations.
\newblock In \emph{The Eleventh International Conference on Learning
  Representations}, 2023.
\newblock URL \url{https://openreview.net/forum?id=Zb6c8A-Fghk}.

\bibitem[Kirillov et~al.(2023)Kirillov, Mintun, Ravi, Mao, Rolland, Gustafson,
  Xiao, Whitehead, Berg, Lo, Doll{\'a}r, and Girshick]{kirillov2023segany}
Alexander Kirillov, Eric Mintun, Nikhila Ravi, Hanzi Mao, Chloe Rolland, Laura
  Gustafson, Tete Xiao, Spencer Whitehead, Alexander~C. Berg, Wan-Yen Lo, Piotr
  Doll{\'a}r, and Ross Girshick.
\newblock Segment anything.
\newblock \emph{arXiv:2304.02643}, 2023.

\bibitem[Koh et~al.(2021)Koh, Sagawa, Marklund, Xie, Zhang, Balsubramani, Hu,
  Yasunaga, Phillips, Gao, Lee, David, Stavness, Guo, Earnshaw, Haque, Beery,
  Leskovec, Kundaje, Pierson, Levine, Finn, and Liang]{koh2021wilds}
Pang~Wei Koh, Shiori Sagawa, Henrik Marklund, Sang~Michael Xie, Marvin Zhang,
  Akshay Balsubramani, Weihua Hu, Michihiro Yasunaga, Richard~Lanas Phillips,
  Irena Gao, Tony Lee, Etienne David, Ian Stavness, Wei Guo, Berton~A.
  Earnshaw, Imran~S. Haque, Sara Beery, Jure Leskovec, Anshul Kundaje, Emma
  Pierson, Sergey Levine, Chelsea Finn, and Percy Liang.
\newblock Wilds: A benchmark of in-the-wild distribution shifts, 2021.

\bibitem[Kuznetsova et~al.(2020)Kuznetsova, Rom, Alldrin, Uijlings, Krasin,
  Pont-Tuset, Kamali, Popov, Malloci, Kolesnikov, Duerig, and
  Ferrari]{OpenImages}
Alina Kuznetsova, Hassan Rom, Neil Alldrin, Jasper Uijlings, Ivan Krasin, Jordi
  Pont-Tuset, Shahab Kamali, Stefan Popov, Matteo Malloci, Alexander
  Kolesnikov, Tom Duerig, and Vittorio Ferrari.
\newblock The open images dataset v4: Unified image classification, object
  detection, and visual relationship detection at scale.
\newblock \emph{IJCV}, 2020.

\bibitem[Li et~al.(2023)Li, Evtimov, Gordo, Hazirbas, Hassner, Ferrer, Xu, and
  Ibrahim]{Li_2023_CVPR_Whac_A_Mole}
Zhiheng Li, Ivan Evtimov, Albert Gordo, Caner Hazirbas, Tal Hassner,
  Cristian~Canton Ferrer, Chenliang Xu, and Mark Ibrahim.
\newblock A whac-a-mole dilemma: Shortcuts come in multiples where mitigating
  one amplifies others.
\newblock In \emph{Proceedings of the IEEE/CVF Conference on Computer Vision
  and Pattern Recognition (CVPR)}, pages 20071--20082, June 2023.

\bibitem[Liu et~al.(2021)Liu, Haghgoo, Chen, Raghunathan, Koh, Sagawa, Liang,
  and Finn]{liu2021just}
Evan~Z Liu, Behzad Haghgoo, Annie~S Chen, Aditi Raghunathan, Pang~Wei Koh,
  Shiori Sagawa, Percy Liang, and Chelsea Finn.
\newblock Just train twice: Improving group robustness without training group
  information.
\newblock In \emph{International Conference on Machine Learning}, pages
  6781--6792. PMLR, 2021.

\bibitem[Liu et~al.(2015)Liu, Luo, Wang, and Tang]{liu2015faceattributes}
Ziwei Liu, Ping Luo, Xiaogang Wang, and Xiaoou Tang.
\newblock Deep learning face attributes in the wild.
\newblock In \emph{Proceedings of International Conference on Computer Vision
  (ICCV)}, December 2015.

\bibitem[Madras and Zemel(2021)]{madras2021identifying}
David Madras and Richard Zemel.
\newblock Identifying and benchmarking natural out-of-context prediction
  problems.
\newblock \emph{Advances in Neural Information Processing Systems},
  34:\penalty0 15344--15358, 2021.

\bibitem[Nam et~al.(2020)Nam, Cha, Ahn, Lee, and Shin]{nam2020learning}
Junhyun Nam, Hyuntak Cha, Sungsoo Ahn, Jaeho Lee, and Jinwoo Shin.
\newblock Learning from failure: Training debiased classifier from biased
  classifier.
\newblock In \emph{Advances in Neural Information Processing Systems}, 2020.

\bibitem[Nam et~al.(2022)Nam, Kim, Lee, and Shin]{nam2022spread}
Junhyun Nam, Jaehyung Kim, Jaeho Lee, and Jinwoo Shin.
\newblock Spread spurious attribute: Improving worst-group accuracy with
  spurious attribute estimation.
\newblock In \emph{International Conference on Learning Representations}, 2022.
\newblock URL \url{https://openreview.net/forum?id=_F9xpOrqyX9}.

\bibitem[Nguyen et~al.(2024)Nguyen, Haddad, Gan, and
  Mirzasoleiman]{nguyen2024make}
Dang Nguyen, Paymon Haddad, Eric Gan, and Baharan Mirzasoleiman.
\newblock Make the most of your data: Changing the training data distribution
  to improve in-distribution generalization performance.
\newblock \emph{arXiv preprint arXiv:2404.17768}, 2024.

\bibitem[Radford et~al.(2021)Radford, Kim, Hallacy, Ramesh, Goh, Agarwal,
  Sastry, Askell, Mishkin, Clark, Krueger, and Sutskever]{radford2021learning}
Alec Radford, Jong~Wook Kim, Chris Hallacy, Aditya Ramesh, Gabriel Goh,
  Sandhini Agarwal, Girish Sastry, Amanda Askell, Pamela Mishkin, Jack Clark,
  Gretchen Krueger, and Ilya Sutskever.
\newblock Learning transferable visual models from natural language
  supervision, 2021.

\bibitem[Rombach et~al.(2022)Rombach, Blattmann, Lorenz, Esser, and
  Ommer]{Rombach_2022_CVPR}
Robin Rombach, Andreas Blattmann, Dominik Lorenz, Patrick Esser, and Bj\"orn
  Ommer.
\newblock High-resolution image synthesis with latent diffusion models.
\newblock In \emph{Proceedings of the IEEE/CVF Conference on Computer Vision
  and Pattern Recognition (CVPR)}, pages 10684--10695, June 2022.

\bibitem[Sagawa* et~al.(2020)Sagawa*, Koh*, Hashimoto, and
  Liang]{sagawa2020distributionally}
Shiori Sagawa*, Pang~Wei Koh*, Tatsunori~B. Hashimoto, and Percy Liang.
\newblock Distributionally robust neural networks.
\newblock In \emph{International Conference on Learning Representations}, 2020.
\newblock URL \url{https://openreview.net/forum?id=ryxGuJrFvS}.

\bibitem[Taori et~al.(2020)Taori, Dave, Shankar, Carlini, Recht, and
  Schmidt]{taori2020measuring}
Rohan Taori, Achal Dave, Vaishaal Shankar, Nicholas Carlini, Benjamin Recht,
  and Ludwig Schmidt.
\newblock Measuring robustness to natural distribution shifts in image
  classification.
\newblock \emph{Advances in Neural Information Processing Systems},
  33:\penalty0 18583--18599, 2020.

\bibitem[Vapnik(1991)]{vapnik}
Vladimir Vapnik.
\newblock Principles of risk minimization for learning theory.
\newblock \emph{Advances in neural information processing systems}, 4, 1991.

\bibitem[Xiao et~al.(2010)Xiao, Hays, Ehinger, Oliva, and Torralba]{sun}
Jianxiong Xiao, James Hays, Krista~A. Ehinger, Aude Oliva, and Antonio
  Torralba.
\newblock Sun database: Large-scale scene recognition from abbey to zoo.
\newblock In \emph{2010 IEEE Computer Society Conference on Computer Vision and
  Pattern Recognition}, pages 3485--3492, 2010.
\newblock \doi{10.1109/CVPR.2010.5539970}.

\bibitem[Yang et~al.(2023{\natexlab{a}})Yang, Nushi, Palangi, and
  Mirzasoleiman]{yang2023mitigating}
Yu~Yang, Besmira Nushi, Hamid Palangi, and Baharan Mirzasoleiman.
\newblock Mitigating spurious correlations in multi-modal models during
  fine-tuning.
\newblock In Andreas Krause, Emma Brunskill, Kyunghyun Cho, Barbara Engelhardt,
  Sivan Sabato, and Jonathan Scarlett, editors, \emph{Proceedings of the 40th
  International Conference on Machine Learning}, volume 202 of
  \emph{Proceedings of Machine Learning Research}, pages 39365--39379. PMLR,
  23--29 Jul 2023{\natexlab{a}}.
\newblock URL \url{https://proceedings.mlr.press/v202/yang23j.html}.

\bibitem[Yang et~al.(2024)Yang, Gan, Karolina~Dziugaite, and
  Mirzasoleiman]{yang2023identifying}
Yu~Yang, Eric Gan, Gintare Karolina~Dziugaite, and Baharan Mirzasoleiman.
\newblock Identifying spurious biases early in training through the lens of
  simplicity bias.
\newblock In Sanjoy Dasgupta, Stephan Mandt, and Yingzhen Li, editors,
  \emph{Proceedings of The 27th International Conference on Artificial
  Intelligence and Statistics}, volume 238 of \emph{Proceedings of Machine
  Learning Research}, pages 2953--2961. PMLR, 02--04 May 2024.
\newblock URL \url{https://proceedings.mlr.press/v238/yang24c.html}.

\bibitem[Yang et~al.(2023{\natexlab{b}})Yang, Zhang, Katabi, and
  Ghassemi]{yang2023change}
Yuzhe Yang, Haoran Zhang, Dina Katabi, and Marzyeh Ghassemi.
\newblock Change is hard: A closer look at subpopulation shift,
  2023{\natexlab{b}}.

\bibitem[Zhang et~al.(2023)Zhang, He, Xu, Yu, Shen, and Cui]{zhang2023nico++}
Xingxuan Zhang, Yue He, Renzhe Xu, Han Yu, Zheyan Shen, and Peng Cui.
\newblock Nico++: Towards better benchmarking for domain generalization.
\newblock In \emph{Proceedings of the IEEE/CVF Conference on Computer Vision
  and Pattern Recognition}, pages 16036--16047, 2023.

\end{thebibliography}
}

\newpage

\onecolumn
\appendix
\section{Existing Methods}\label{appendix:methods}

Several methods have been proposed to prevent models from exploiting spurious correlations and improve worst group error. 
If group labels are available, group robust optimization or sampling methods upweight or upsample the minority groups to achieve a similar accuracy on all groups. 
If group labels are not available, existing methods first infer groups of the training data and then train the model using robust optimization or sampling techniques with the inferred group labels. \looseness=-1

\subsection{With Group Labeled Training Data}



\textbf{Group Balancing} samples every mini-batch to have equal examples from each group.

\textbf{GroupDRO (GDRO)} leverages group information to sample group-balanced batches of training data, and minimizes the empirical worst-group training loss:
\begin{equation}
\w\in \argmin \max_{g \in \mathcal{G}}  \E_{(\x_i,y_i)\in g}[l(f(\w,\x_i),y_i)].
\end{equation}
GroupDRO solves the above optimization problem by maintaining a weight $q_g$ for each group $g$ and weighting the loss of examples in group $g$ by $q_g$. Stochastic gradient descent on parameter $\w$ is interleaved with gradient ascent on the weights $q_g$. GDRO uses a very small tunable learning rate and a large regularizer to achieve a satisfactory performance.

\textbf{PDE} is a two-stage training algorithm designed to mitigate spurious correlations by progressively expanding the training data. In the warm-up stage, a small, balanced subset of data is used to prevent the model from learning spurious features. In the expansion stage, small random subsets of the remaining data are incrementally added, leveraging the momentum from the warm-up stage to continue learning core features effectively.

\subsection{Without Group Labeled Data}

In absence of group labels, methods typically infer groups, often using a reference model $f_{ref}$ trained with ERM, and then leverage the inferred groups to train a robust model using sampling or GDRO. 

\textbf{Just Train Twice (JTT)} \cite{liu2021just} first trains a reference model $f_{ref}$ using ERM for $T_r$ number of epochs. Then, it identifies 
the minority group as examples that are misclassified 
by the reference model.
JTT then upsamples the misclassified examples $S_r$ times and trains another model using ERM on the upsampled dataset. $T_r$ and $S_r$ are tuned to achieve the optimal performance.

\textbf{Environment Inference for Invariant Learning (EIIL)} 
EIIL consists of two stages: environment (Group) inference (EI) and invariant learning (IL). In the EI stage, EIIL aims to infer the worst-case groups (environments) using a reference model $f_{ref}$ that has been trained with ERM by optimizing a soft-group assignment $\pmb{q}$ to maximize the following objective: 
\begin{align}
    C^{EI}(f_{ref},\pmb{q}) = \| \nabla_{\bar{\w}}\tilde{R}^e(\bar{\w} \cdot f_{ref}, \pmb{q})\|.
\quad \\ \text{s.t.} \quad
    \tilde{R}^e(f_{ref},\pmb{q}) = \frac{1}{N} \sum_i q_i(e)l(f_{ref}(\x_i),y_i).
\end{align}
where $\bar{\w}$ is the all-ones vector. Intuitively, the group assignment $\pmb{q}$ is optimized to find examples that are most sensitive to small changes in the reference model $f_{ref}$'s outputs.
%
In the IL stage, EIIL uses GroupDRO with the inferred groups to train a robust model.

\textbf{Separate Early and Re-sample (SPARE)} SPARE \cite{yang2023identifying} 
proved that in the presence of strong spurious correlations, the outputs of a neural network trained with ERM are mainly determined by the spurious features, early-in-training. Thus, it infers groups by clustering the reference model $f_{ref}$'s output on each class, where $f_{ref}$ is trained for a few $T_r$ epochs. 
The inferred groups (clusters) are then used to train a model with importance sampling based on the cluster sizes, to balance the groups. Number of clusters 
and $T_r$ are tuned to achieve best performance.



\subsection{Using the Group-Labeled Validation Set}

\textbf{Spread Spurious Attributes (SSA)} SSA \citep{nam2022spread} leverages the group labeled validation data to 
train a group label predictor
in a semi-supervised manner. Given a training example $\x_i$, let $g$ denotes its 
(unknown) group label, $\hat{p}(\cdot|\x_i)$ denotes the model's predicted group label distribution and $\hat{g}\coloneqq\arg\max_{g\in\mathcal{A}}\hat{p}(g|\x_i)$ denotes the predicted group label. SSA minimizes the following loss function: \looseness=-1
\begin{align}
    \nonumber
    \mathcal{L} = & \hat{\mathbbm{E}}_{\text{group labeled}}[\CE(\hat{p}(\cdot|\x), g)] \\ + & \hat{\mathbbm{E}}_{\text{group-unlabeled}}[\mathbbm{1}_{\max_{g\in\mathcal{A}}\hat{p}(g|\x)\geq \tau} \CE(\hat{p}(\cdot|\x), \hat{g})],
\end{align}
where CE is the cross entropy loss. 
Effectively, in addition to fitting the available group labels (validation set), SSA uses the group unlabeled examples with confident predictions (larger than a threshold $\tau$) as additional group labeled data. 
Moreover, SSA partitions the group unlabeled data into multiple splits and trains a group label predictor for each left-out split by using the other splits as the group unlabeled training data. 
The threshold $\tau$ and the number of splits are both tuned to achieve optimal performance. \looseness=-1

\textbf{Deep Feature Reweighting (DFR)} DFR \cite{kirichenko2023last} 
argues that a model trained with ERM captures both core and spurious features in its last hidden layer, even though it may exhibit spurious correlations in its predictions. Motivated by this, DFR retrains the last linear layer of the model on a group-balanced validation data, while keeping earlier layers frozen. This enables the model to adjust the weights assigned to the features in the penultimate layer. 
DFR trains multiple linear models with tunable  $\ell_1$ regularization on randomly sampled group-balanced validation data and averages their weights. 

\newpage
\section{\animals Details}
\label{appendix:spuco_animals}

\subsection{Dataset Details}

\subsubsection{Grouping of Fine-Grained ImageNet Classes for SpuCoAnimals Classes}

\textbf{Landbirds and Waterbirds}. Landbirds are birds that primarily inhabit terrestrial environments, while waterbirds are birds that primarily inhabit aquatic or semi-aquatic habitats. This is identical to the grouping of birds used by Waterbirds \cite{sagawa2020distributionally}. 

\texttt{landbirds = [rooster, hen, ostrich, brambling, goldfinch, house finch, junco, indigo bunting, American robin, bulbul, jay, magpie, chickadee, American dipper, kite (bird of prey), bald eagle, vulture, great grey owl, black grouse, ptarmigan, ruffed grouse, prairie grouse, peafowl, quail, partridge, african grey parrot, macaw, sulphur-crested cockatoo, lorikeet, coucal, bee eater, hornbill, hummingbird, jacamar, toucan]}

\texttt{waterbirds = [duck, red-breasted merganser, goose, black swan, white stork, black stork, spoonbill, flamingo, little blue heron, great egret, bittern bird, crane bird, limpkin, common gallinule, American coot, bustard, ruddy turnstone, dunlin, common redshank, dowitcher, oystercatcher, pelican, king penguin, albatross]}

\textbf{Small Dogs and Big Dogs}. The two lists of dogs correspond to categorization based on their respective sizes. The first list consists of small dog breeds commonly referred to as "toy" or "toy terrier" breeds. These dogs are typically small in size and often kept as companion pets. Examples include Chihuahua, Maltese, Shih Tzu, and Yorkshire Terrier. The second list consists of a variety of dog breeds, including medium-sized and large-sized breeds. These breeds are specifically not classified as toy breeds. Examples from this list include Labrador Retriever, German Shepherd Dog, Rottweiler, Great Dane, Alaskan Malamute, and Siberian Husky. These breeds are often known for their working abilities, guarding skills, or other specific purposes.

\texttt{small dog breeds = [Chihuahua, Japanese Chin, Maltese, Pekingese, Shih Tzu, King Charles Spaniel, Papillon, toy terrier, Italian Greyhound, Whippet, Ibizan Hound, Norwegian Elkhound, Yorkshire Terrier, Norfolk Terrier, Norwich Terrier, Wire Fox Terrier, Lakeland Terrier, Sealyham Terrier, Cairn Terrier, Australian Terrier, Dandie Dinmont Terrier, Boston Terrier, Miniature Schnauzer, Scottish Terrier, Tibetan Terrier, Australian Silky Terrier, West Highland White Terrier, Lhasa Apso, Soft-coated Wheaten Terrier, Australian Kelpie, Shetland Sheepdog, Pembroke Welsh Corgi, Cardigan Welsh Corgi, Toy Poodle, Miniature Poodle, Mexican hairless dog (xoloitzcuintli)]}

\texttt{big dog breeds = [Rhodesian Ridgeback, Afghan Hound, Basset Hound, Beagle, Bloodhound, Bluetick Coonhound, Black and Tan Coonhound, Treeing Walker Coonhound, English foxhound, Redbone Coonhound, borzoi, Irish Wolfhound, Otterhound, Saluki, Scottish Deerhound, Weimaraner, Staffordshire Bull Terrier, American Staffordshire Terrier, Bedlington Terrier, Border Terrier, Kerry Blue Terrier, Irish Terrier, Flat-Coated Retriever, Curly-coated Retriever, Golden Retriever, Labrador Retriever, Chesapeake Bay Retriever, German Shorthaired Pointer, Vizsla, English Setter, Irish Setter, Gordon Setter, Brittany dog, Clumber Spaniel, English Springer Spaniel, Welsh Springer Spaniel, Cocker Spaniel, Sussex Spaniel, Irish Water Spaniel, Kuvasz, Schipperke, Groenendael dog, Malinois, Briard, Komondor, Old English Sheepdog, collie, Border Collie, Bouvier des Flandres dog, Rottweiler, German Shepherd Dog, Dobermann, Greater Swiss Mountain Dog, Bernese Mountain Dog, Appenzeller Sennenhund, Entlebucher Sennenhund, Boxer, Bullmastiff, Tibetan Mastiff, French Bulldog, Great Dane, St. Bernard, Alaskan Malamute, Siberian Husky, Leonberger, Newfoundland dog, Great Pyrenees dog, Samoyed, Chow Chow, Keeshond, Dalmatian, Affenpinscher, Basenji, pug]}

\subsubsection{Statistics about Data}

\begin{table}[h]
\caption{Number of Examples per Group in Train, Validation and Test Sets of \animals}
\resizebox{\columnwidth}{!}{%
\begin{tabular}{|l|ll|ll|ll|ll|}
\toprule
Data Split          & \multicolumn{2}{c|}{\textbf{Landbirds}}     & \multicolumn{2}{c|}{\textbf{Waterbirds}}   & \multicolumn{2}{c|}{\textbf{Small Dogs}}       & \multicolumn{2}{c|}{\textbf{Big Dogs}}  \\
           & \multicolumn{1}{l}{Land}  & Water & \multicolumn{1}{l|}{Land} & Water & \multicolumn{1}{l|}{Indoor} & Outdoor & \multicolumn{1}{l|}{Indoor} & Outdoor \\ \midrule \midrule
Train      & \multicolumn{1}{l|}{10000} & 500   & \multicolumn{1}{l|}{500}  & 10000 & \multicolumn{1}{l|}{10000}  & 500     & \multicolumn{1}{l|}{500}    & 10000   \\ 
Validation & \multicolumn{1}{l|}{500}   & 25    & \multicolumn{1}{l|}{25}   & 500   & \multicolumn{1}{l|}{500}    & 25      & \multicolumn{1}{l|}{25}     & 500     \\ 
Test      & \multicolumn{1}{l|}{500}   & 500   & \multicolumn{1}{l|}{500}  & 500   & \multicolumn{1}{l|}{500}    & 500     & \multicolumn{1}{l|}{500}    & 500 \\   
\bottomrule
\end{tabular}
}
\end{table}

\textbf{Variety in Spurious Features}

Here, we show how many examples for each spurious environment correspond to each of the prompts used to identify the environments. This illustrates the diversity in spurious features in SpuCoAnimals, which as shown in \cite{yang2023change} determines the effectiveness of methods to improve worst-group accuracy. 

For spurious feature = “outdoor”: 
Big Dog Outdoor: (grass, 9222), (park, 734), (road, 44)
Small Dog Outdoor: (grass, 472), (park, 25), (road, 3)

For spurious feature = “indoor”: 
Big Dog Indoor: (couch, 5408), (floor, 3252), (bed, 1340)
Small Dog Indoor: (couch, 266), (floor, 196), (bed, 38)

For spurious feature = “land”: 
Landbirds Land: (grass, 7441), (tree, 1395), (tree, 1164)
Waterbirds Land: (grass, 469), (forest, 18), (tree, 13)

For spurious feature = “water”: 
Landbirds Water: (river, 218), (sea, 179), (lake, 103)
Waterbirds Water: (lake, 5367), (sea, 3640), (river, 989), (ocean, 4)

\newpage
\subsubsection{Examples}

\begin{figure}[h]
    \begin{subfigure}{\textwidth}
    \includegraphics[width=0.24\textwidth]{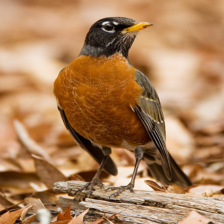} 
    \includegraphics[width=0.24\textwidth]{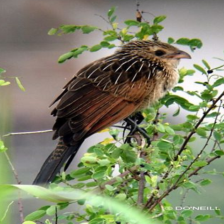} 
    \includegraphics[width=0.24\textwidth]{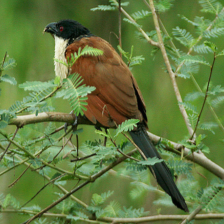} 
    \includegraphics[width=0.24\textwidth]{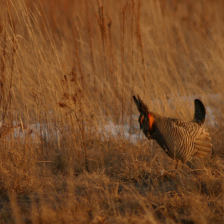} 
    \caption{Landbirds on Land}
    \end{subfigure}

    \begin{subfigure}{\textwidth}
    \includegraphics[width=0.24\textwidth]{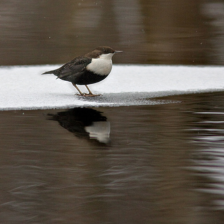} 
    \includegraphics[width=0.24\textwidth]{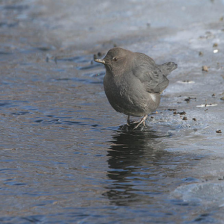} 
    \includegraphics[width=0.24\textwidth]{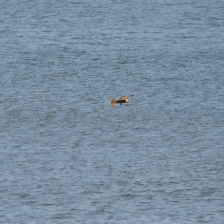} 
    \includegraphics[width=0.24\textwidth]{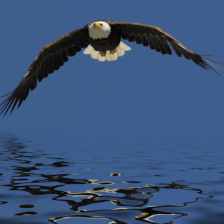} 
    \caption{Landbirds on Water}
    \end{subfigure}

    \begin{subfigure}{\textwidth}
    \includegraphics[width=0.24\textwidth]{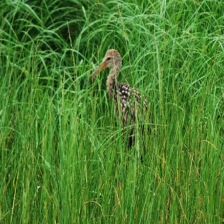} 
    \includegraphics[width=0.24\textwidth]{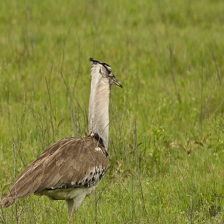} 
    \includegraphics[width=0.24\textwidth]{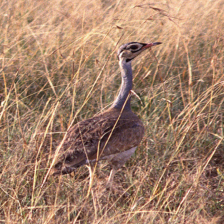} 
    \includegraphics[width=0.24\textwidth]{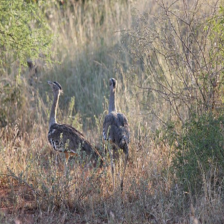} 
    \caption{Waterbirds on Land}
    \end{subfigure}

    \begin{subfigure}{\textwidth}
    \includegraphics[width=0.24\textwidth]{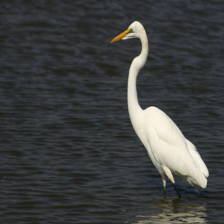} 
    \includegraphics[width=0.24\textwidth]{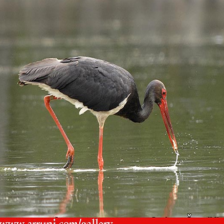} 
    \includegraphics[width=0.24\textwidth]{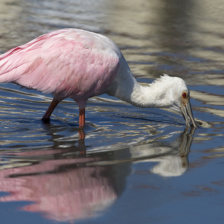} 
    \includegraphics[width=0.24\textwidth]{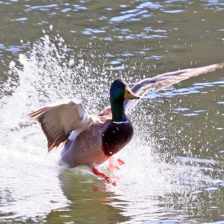} 
    \caption{Waterbirds on Water}
    \end{subfigure}
    \caption{Examples from Bird Groups}
    \label{fig:app_examples_birds}
    \vspace{-5mm}
\end{figure}

\begin{figure*}[h]
    \begin{subfigure}{\textwidth}
    \includegraphics[width=0.24\textwidth]{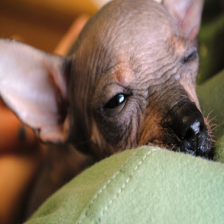} 
    \includegraphics[width=0.24\textwidth]{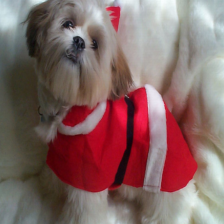} 
    \includegraphics[width=0.24\textwidth]{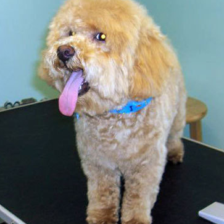} 
    \includegraphics[width=0.24\textwidth]{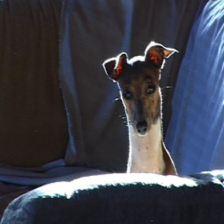} 
    \caption{Small Dogs Indoor}
    \end{subfigure}

    \begin{subfigure}{\textwidth}
    \includegraphics[width=0.24\textwidth]{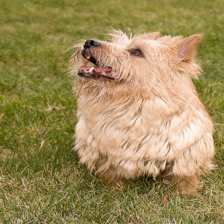} 
    \includegraphics[width=0.24\textwidth]{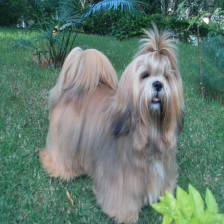} 
    \includegraphics[width=0.24\textwidth]{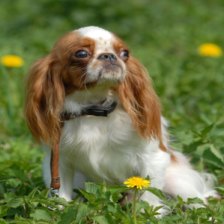} 
    \includegraphics[width=0.24\textwidth]{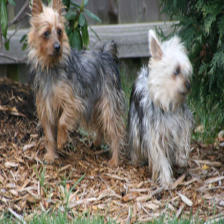} 
    \caption{Small Dogs Outdoor}
    \end{subfigure}

    \begin{subfigure}{\textwidth}
    \includegraphics[width=0.24\textwidth]{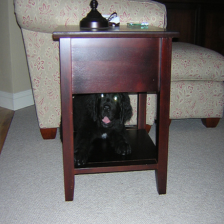} 
    \includegraphics[width=0.24\textwidth]{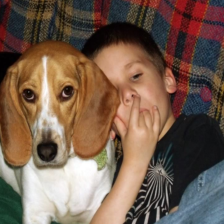} 
    \includegraphics[width=0.24\textwidth]{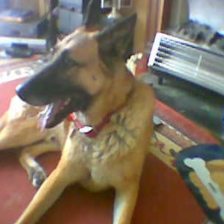} 
    \includegraphics[width=0.24\textwidth]{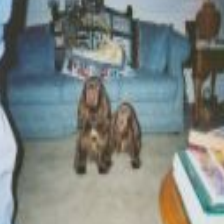} 
    \caption{Big Dogs Indoor}
    \end{subfigure}

    \begin{subfigure}{\textwidth}
    \includegraphics[width=0.24\textwidth]{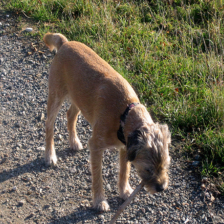} 
    \includegraphics[width=0.24\textwidth]{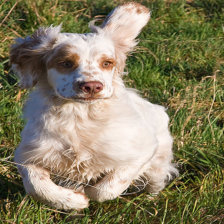} 
    \includegraphics[width=0.24\textwidth]{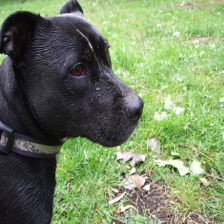} 
    \includegraphics[width=0.24\textwidth]{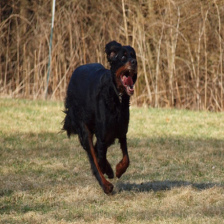} 
    \caption{Big Dogs Outdoor}
    \end{subfigure}
    \caption{Examples from Dog Groups}
    \label{fig:app_examples_dogs}
\end{figure*}

Fig. \ref{fig:app_examples_birds} and Fig. \ref{fig:app_examples_dogs} show examples from each group of examples from birds and dogs respectively. 

\subsection{Ablations on SpuCoSun}\label{apdx: ablation_sun}

To demonstrate the learnability of the core feature on \sun, we construct two additional versions of the dataset where no spurious correlation exists: (1) Masked-Spurious, where we mask out the spurious features, and (2) Core-Only, where we do not add the spurious feature to the core, e.g., just the corresponding original image from the SUN397 dataset. The accuracy achieved by ERM on these modified datasets should serve as the upper bound for our target performance, as this represents the scenario where no spurious correlation issues are present. In Table \ref{tab:no_mask}, we show the per-group accuracy achieved by ERM on both datasets. We see that the worst-group accuracy here is higher than those presented for \sun in Table \ref{tab:all_tab}, indicating significant room for improvement for existing methods, including those with known group information.

\begin{table}[!t]
    \caption{Accuracy on each group for models trained and evaluated on the modified versions of \sun where no spurious correlation exists. The model is trained using ERM. Here, Group ID is in the format of (Class ID, Spurious Attribute ID).}
    \label{tab:no_mask}
    \centering
    \begin{tabular}{|c|c|c|}
    \hline
        Group ID & Masked-Spurious & Core-Only \\
        \hline
        (0, 0) & 73.6 & 82.8 \\
        (0, 1) & 73.3 & 84.1 \\
        (0, 2) & 71.3 & 85.3 \\
        (0, 3) & 72.1 & 83.7 \\
        (1, 0) & 76.5 & 92.4 \\
        (1, 1) & 82.0 & 94.4 \\
        (1, 2) & 72.1 & 88.5 \\
        (1, 3) & 76.1 & 92.8 \\
        (2, 0) & 75.7 & 84.5 \\
        (2, 1) & 76.5 & 86.5 \\
        (2, 2) & 76.4 & 83.6 \\
        (2, 3) & 73.7 & 85.7 \\
        (3, 0) & 76.9 & 87.3 \\
        (3, 1) & 79.7 & 84.9 \\
        (3, 2) & 75.7 & 87.6 \\
        (3, 3) & 83.2 & 84.8 \\
        \hline
    \end{tabular}
\end{table}

\newpage
\section{SpuCoSun} \label{appendix:spuco_sun}

\subsection{Construction of \sun}

The \sun dataset was constructed through a series of steps to ensure a diverse and challenging dataset for machine learning experiments. Below are the detailed steps involved in its creation:

\begin{enumerate}
    \item \textbf{Selection of Superclasses from SUN397:}
    \begin{itemize}
        \item Four superclasses were chosen from the SUN397 dataset: Recreational, Residential, Cultural, and Infrastructure.
        \item These superclasses formed the basis for the classes in the \sun dataset.
    \end{itemize}

    \item \textbf{Incorporation of Spurious Features:}
    \begin{itemize}
        \item To add spurious features, we introduced co-occurring objects from the OpenImagesV7 dataset.
        \item The chosen objects were from specific categories: sports equipment (basketball, golf ball, tennis ball), fruits \& vegetables (pumpkin, watermelon, broccoli), baked goods (muffin, bagel, pretzel), and containers (waste container, can, barrel).
    \end{itemize}

    \item \textbf{Variation in Spurious Features:}
    \begin{itemize}
        \item For \suneasy, the spurious co-occurring object was sampled from only one subclass.
        \item For \sun, the spurious feature's variation was increased by sampling the spurious co-occurring object from all three subclasses, making the learning task more challenging.
    \end{itemize}

    \item \textbf{Generation of Co-occurring Objects:}
    \begin{itemize}
        \item We used a Text-To-Image Latent Diffusion Model \cite{Rombach_2022_CVPR} to generate the co-occurring objects.
        \item The prompt used for generating these objects was \textit{"a \textbf{classname }on black background"}.
    \end{itemize}

    \item \textbf{Image Resizing:}
    \begin{itemize}
        \item The background image, representing the class feature, was resized to (224,224) pixels.
        \item This resizing was done for all selected images from SUN397.
    \end{itemize}

    \item \textbf{Segmentation of Co-occurring Objects:}
    \begin{itemize}
        \item The SegmentAnythingModel \cite{kirillov2023segany} was utilized to obtain a fine-grained mask of the co-occurring object generated by the diffusion model.
    \end{itemize}

    \item \textbf{Resizing Co-occurring Objects:}
    \begin{itemize}
        \item The segmented co-occurring object was resized, preserving its aspect ratio, to fit within the central (112,112) pixels of the (224,224) background image.
    \end{itemize}

    \item \textbf{Validation of Dataset without Spurious Features:}
    \begin{itemize}
        \item To confirm that the spurious feature's masking does not render the class unlearnable, we created a version of the dataset without the spurious feature.
        \item In this version, the central (112,112) pixels were masked out, and a CLIP pre-trained ResNet-50 model was trained with ERM on this modified dataset.
        \item The model achieved an accuracy of 76.8\%, indicating that the task remains solvable even with the spurious feature occlusion.
    \end{itemize}
\end{enumerate}

These steps collectively ensured the \sun dataset was robust and suitable for evaluating the influence of spurious features in machine learning tasks.

\newpage
\subsection{Empirically Confirming that Spurious Feature of \textsc{\suneasy} is easier than that of \textsc{\sun}}

Figure \ref{fig:spucosun_easiness_evidence} shows that the spurious feature of \suneasy is learned faster than the spurious feature of \sun. 

\begin{figure}[h]
    \centering
    \includegraphics[width=0.5\textwidth]{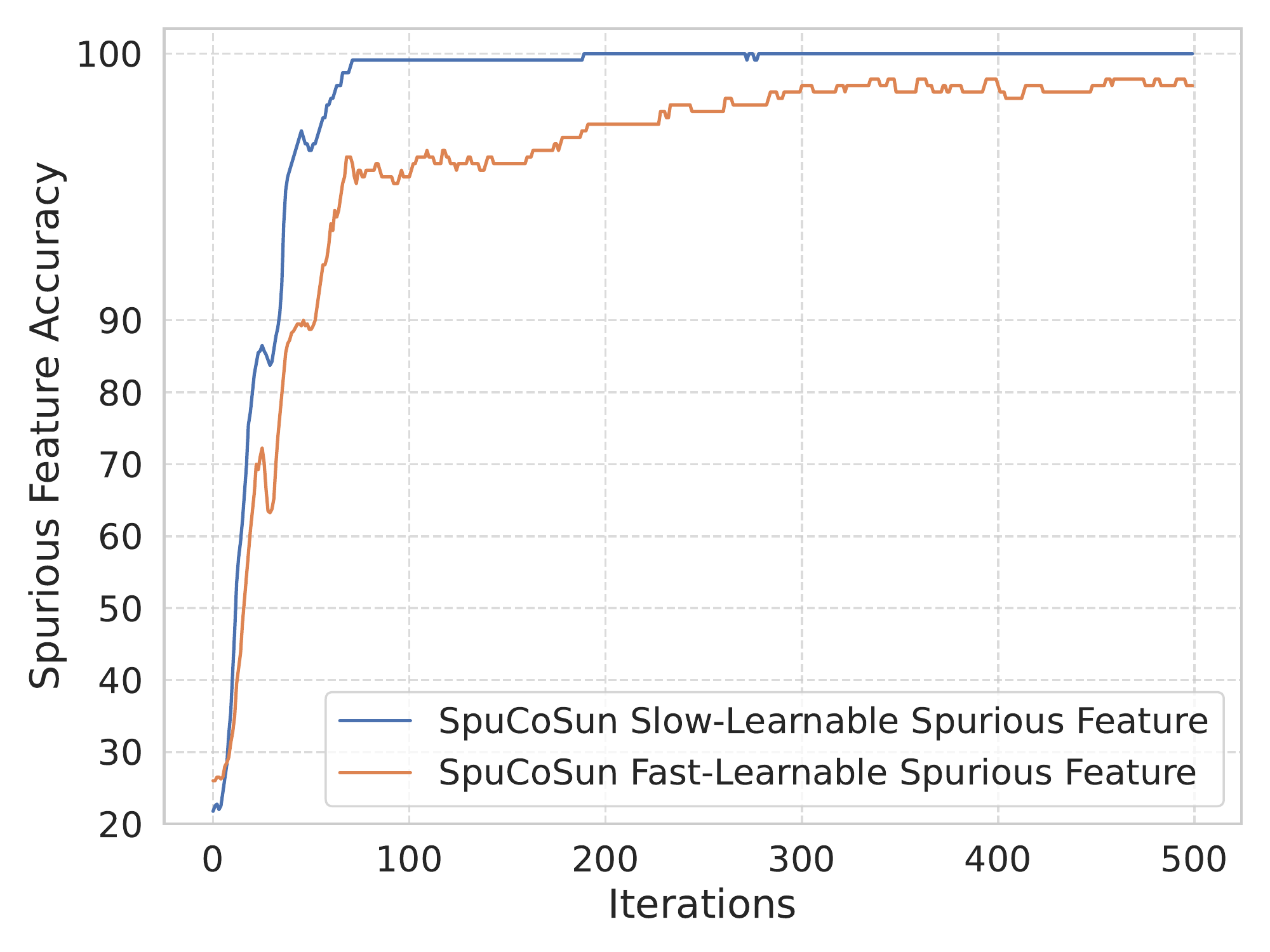}
    \caption{Comparing Average Accuracies of Majority and Minority Groups on \textsc{\suneasy} v/s \textsc{\sun}}
    \label{fig:spucosun_easiness_evidence}
\end{figure}

\newpage
\section{Optimal Hyperparameters (used for results reported in Table \ref{tab:all_tab} and \ref{tab:spu_diff_tab})} \label{appendix:hparams_main}

In this section, we present the optimal hyperparameters found via our model selection to report the results in Table \ref{tab:all_tab}. For existing datasets i.e. \water, \celeba, \cars, we follow previous work and use a ResNet-50, pre-trained on ImageNet. For the new datasets proposed in this work i.e. \animals, \sun (Easy) and \sun (Hard), we use a ResNet-50, pretrained with CLIP \cite{radford2021learning} and only train the final projection layer as in \cite{yang2023mitigating}. For all experiments, we use batch size of 128 as is standard on previous datasets.

\begin{table}[h]
\centering
\caption{Number of Epochs for Robust Training on Each Dataset}
\begin{tabular}{|l|ccc|}
\hline
 & \water & \celeba & \cars \\
\hline
\hline
Epochs & 300 & 50 & 300 \\
\hline
\hline
 & \animals & \sun (Hard) & \sun (Easy) \\
\hline
\hline
Epochs & 100 & 40 & 40 \\
\hline
\end{tabular}
\label{tab:num_epochs_dataset}
\end{table}

\begin{table}[h]
\centering
\caption{SPARE}
\begin{tabular}{|l|ccc|}
\hline
 & \water & \celeba & \cars \\
\hline
\hline
LR & 1e-3 & 1e-5 & 1e-4 \\
WD & 1e-4 & 1e-0 & 1e-1 \\
Infer epoch & 2 & 1 & 1 \\
N clusters & 2 & 2 & 4 \\
Upsample power & 3 & 2 & 2 \\
\hline
\hline
 & \animals & \sun (Hard) & \sun (Easy) \\
\hline
\hline
LR & 1e-4 & 1e-3 & 1e-4 \\
WD & 1e-2 & 1e-2 & 1e-1 \\
Infer epoch & 2 & 1 & 2 \\
N clusters & 2 & 4 & 4 \\
Upsample power & 2 & 2 & 1 \\
\hline
\end{tabular}
\label{tab:hyper-spare}
\end{table}

\begin{table}[h]
\centering
\caption{PDE}
\begin{tabular}{|l|ccc|}
\hline
 & \water & \celeba & \cars \\
\hline
\hline
LR & 1e-2 & 1e-2 & 1e-3 \\
WD & 1e-2 & 1e-4 & 1e-1 \\
Warmup length & 140 & 16 & 75 \\
Expand size & 10 & 50 & 10 \\
Expand interval & 10 & 10 & 30 \\
Subsample min & - & - & 150 \\
\hline
\hline
 & \animals & \sun (Hard) & \sun (Easy) \\
\hline
\hline
LR & 1e-3 & 1e-2 & 1e-2 \\
WD & 1e-3 & 1e-4 & 1e-3 \\
Warmup length & 20 & 20 & 20 \\
Expand size & 10 & 50 & 10 \\
Expand interval & 2 & 2 & 2 \\
Subsample min & - & - & - \\
\hline
\end{tabular}
\label{tab:hyper-pde}
\end{table}

\begin{table}[h]
\centering
\caption{ERM}
\begin{tabular}{|l|ccc|}
\hline
 & \water & \celeba & \cars \\
\hline
\hline
LR & 1e-3 & 1e-4 & 1e-4 \\
WD & 1e-4 & 1e-1 & 1e-1 \\
\hline
\hline
 & \animals & \sun (Hard) & \sun (Easy) \\
\hline
\hline
LR & 1e-3 & 1e-5 & 1e-5 \\
WD & 1e-4 & 1e-4 & 1e-4 \\
\hline
\end{tabular}
\label{tab:hyper-erm}
\end{table}

\begin{table}[h]
\centering
\caption{GDRO}
\begin{tabular}{|l|ccc|}
\hline
 & \water & \celeba & \cars \\
\hline
\hline
LR & 1e-5 & 1e-4 & 1e-4 \\
WD & 1 & 1 & 1 \\
\hline
\hline
 & \animals & \sun (Hard) & \sun (Easy) \\
\hline
\hline
LR & 1e-5 & 1e-5 & 1e-5 \\
WD & 1e-4 & 1 & 1 \\
\hline
\end{tabular}
\label{tab:hyper-gdro}
\end{table}

\begin{table}[h]
\centering
\caption{GB}
\begin{tabular}{|l|ccc|}
\hline
 & \water & \celeba & \cars \\
\hline
\hline
LR & 1e-5 & 1e-4 & 1e-4 \\
WD & 1 & 1 & 1 \\
\hline
\hline
 & \animals & \sun (Hard) & \sun (Easy) \\
\hline
\hline
LR & 1e-5 & 1e-5 & 1e-5 \\
WD & 1e-4 & 1 & 1 \\
\hline
\end{tabular}
\label{tab:hyper-gb}
\end{table}

\begin{table}[h]
\centering
\caption{JTT}
\begin{tabular}{|l|ccc|}
\hline
 & \water & \celeba & \cars \\
\hline
\hline
LR & 1e-5 & 1e-5 & 1e-4 \\
WD & 1 & 1e-1 & 1e-1 \\
Infer epoch & 60 &1  & 2 \\
Upsampling Power & 100 & 50 & 50 \\
\hline
\hline
 & \animals & \sun (Hard) & \sun (Easy) \\
\hline
\hline
LR & 1e-3 & 1e-3 & 1e-3 \\
WD & & 1e-5 & 1e-5 \\
Infer epoch & 1 & 1 & 2 \\
Upsampling Power & 100 & 100 & 100 \\
\hline
\end{tabular}
\label{tab:hyper-jtt}
\end{table}

\begin{table}[h]
\centering
\caption{EIIL}
\begin{tabular}{|l|ccc|}
\hline
 & \water & \celeba & \cars \\
\hline
\hline
ERM LR & 1e-3 & 1e-4 & 1e-4 \\
ERM WD & 1e-4 & 1e-4 & 1e-1 \\
GDRO LR & 1e-3 & 1e-5 & 1e-5 \\
GDRO WD & 1e-4 & 1e-1 & 1 \\
Infer epoch & 1 & 1 & 2 \\
EIIL LR & 1e-2 & 1e-3 & 1e-3 \\
EIIL steps & 20000 & 10000 & 20000 \\
\hline
\hline
 & \animals & \sun (Hard) & \sun (Easy) \\
\hline
\hline
ERM LR & 1e-3 & 1e-3 & 1e-3 \\
ERM WD & 1e-4 & 1e-3 & 1e-3 \\
GDRO LR & 1e-4 & 1e-4 & 1e-4 \\
GDRO WD & 1e-5 & 1e-4 & 1e-3 \\
Infer epoch & 2 & 2 & 1 \\
EIIL LR & 1e-3 & 1e-2 & 1e-2 \\
EIIL steps & 10000 & 10000 & 20000 \\
\hline
\end{tabular}
\label{tab:hyper-eiil}
\end{table}

\begin{table}[h]
\centering
\caption{DFR}
\begin{tabular}{|l|ccc|}
\hline
 & \water & \celeba & \cars \\
\hline
\hline
ERM LR & 1e-3 & 1e-4 & 1e-3 \\
ERM WD & 1e-3 & 1e-1 & 1e-1 \\
\hline
\hline
 & \animals & \sun (Hard) & \sun (Easy) \\
\hline
\hline
ERM LR & 1e-3 & 1e-3 & 1e-3 \\
ERM WD & 1e-4 & 1e-4 & 1e-4 \\
\hline
\end{tabular}
\label{tab:hyper-dfr}
\end{table}

\begin{table}[h]
\centering
\caption{SSA}
\begin{tabular}{|l|ccc|}
\hline
 & \water & \celeba & \cars \\
\hline
\hline
SSL LR & 1e-4 & 1e-5 & 1e-4 \\
SSL WD & 1e-1 & 1e-1 & 1 \\
GDRO LR & 1e-4 & 1e-4 & 1e-3 \\
GDRO WD & 1e-1 & 1 & 1e-4 \\
\hline
\hline
 & \animals & \sun (Hard) & \sun (Easy) \\
\hline
\hline
SSL LR & 1e-4 & 1e-2 & 1e-4 \\
SSL WD & 1e-3 & 1e-2 & 1e-5 \\
GDRO LR & 1e-3 & 1e-3 & 1e-4 \\
GDRO WD & 1e-4 & 1e-2 & 1e-3 \\
\hline
\end{tabular}
\label{tab:hyper-ssa}
\end{table}

\newpage
\section{Model Selection} \label{apdx: selection}

\subsection{Group Inference Evaluation Metrics}  \label{apdx: metrics}

For the group inference metrics, we consider each class as having two groups:
1) examples that have the spurious feature correlated with this class in the training data (the majority group in the training data, henceforth referred to as the \textit{majority} group), and 
2) examples that do not have the spurious feature correlated with this class in the training data (the minority group(s) in the training data, henceforth referred to as the \textit{minority} group).

Note that in datasets such as \sun and \cars, there are multiple \textit{groups} that do not have the spurious feature correlated with the class. In these cases, the \textit{minority} group refers to the union of all these groups.

Having defined two groups per class, we evaluate group inference metrics by assessing how often they correctly identify whether an example belongs to the \textit{majority} group or the \textit{minority} group. Here, belonging to the \textit{minority} group is considered the positive label, and precision and recall are defined with respect to this label.

Since we evaluate accuracy, precision, and recall per class, our analysis includes both average precision/recall and minimum precision/recall per class. For accuracy, we only report the overall accuracy across all classes, rather than considering minimum accuracy.

We now go into specific details about how we compute group inference metrics for each method considered in this paper. Our approach is general and can be applied to any group inference method. 

\textbf{JTT} For JTT, we classify the examples of the validation set, and those misclassified are determined as the error set. We treat this as the \textit{minority} group and the rest as the \textit{majority} group. We then partition these groups using the provided class labels, allowing us to evaluate our group inference metrics.

\textbf{EIIL} For EIIL, we run the group inference algorithm on the validation set. Since the groups are symmetric for EIIL, on a group-balanced validation set, either group could be the \textit{majority} or \textit{minority}. Therefore, we assign the \textit{majority} and \textit{minority} labels to both groups and select the assignment with the higher accuracy to ensure our labeling is correct. We then use this assignment to compute the accuracy, recall, and precision metrics.

\textbf{SPARE} For SPARE, we run the clustering algorithm on the validation set to determine validation set groups. Similar to EIIL, on a group-balanced validation set, any of the groups could correspond to the \textit{majority}. Therefore, we first consider each group of a class as the \textit{majority}, then merge all other groups into the \textit{minority} and evaluate accuracy. We select the \textit{majority} assignment that achieves the highest accuracy. This process is repeated for each class since SPARE infers groups per class. We then use this assignment to compute the accuracy, recall, and precision metrics.


\subsection{Hyperparameter Ranges for Different Methods on \animals}
\label{sec:hyper_range}

The following commands and the hyperparameter ranges for various methods are provided below:

\begin{itemize}
    \item \textbf{ERM:}
    \begin{itemize}
        \item Command: \texttt{python spuco\_animals\_erm.py}
        \item Hyperparameters:
        \begin{itemize}
            \item \texttt{lr} = [1e-3, 1e-4, 1e-5]
            \item \texttt{weight\_decay} = [1e-2, 1e-3, 1e-4]
        \end{itemize}
    \end{itemize}

    \item \textbf{Group Balancing:}
    \begin{itemize}
        \item Command: \texttt{python spuco\_animals\_gb.py}
        \item Hyperparameters:
        \begin{itemize}
            \item \texttt{lr} = [1e-3, 1e-4, 1e-5]
            \item \texttt{weight\_decay} = [1e-2, 1e-3, 1e-4]
        \end{itemize}
    \end{itemize}

    \item \textbf{GroupDRO:}
    \begin{itemize}
        \item Command: \texttt{python spuco\_animals\_gdro.py}
        \item Hyperparameters:
        \begin{itemize}
            \item \texttt{lr} = [1e-3, 1e-4, 1e-5]
            \item \texttt{weight\_decay} = [1e-2, 1e-3, 1e-4]
        \end{itemize}
    \end{itemize}
    
    \item \textbf{EIIL:}
    \begin{itemize}
        \item Command: \texttt{python quickstart/spuco\_animals/spuco\_animals\_eiil.py}
        \item Hyperparameters:
        \begin{itemize}
            \item \texttt{erm\_lr} = [1e-3, 1e-4]
            \item \texttt{erm\_weight\_decay} = [1e-3, 1e-4]
            \item \texttt{gdro\_lr} = [1e-5, 1e-4]
            \item \texttt{gdro\_weight\_decay} = [1e-1, 1e-0]
            \item \texttt{infer\_num\_epochs} = [1, 2]
            \item \texttt{eiil\_num\_steps} = [10000, 20000]
            \item \texttt{eiil\_lr} = [1e-2, 1e-3]
        \end{itemize}
    \end{itemize}

    \item \textbf{SPARE:}
    \begin{itemize}
        \item Command: \texttt{python quickstart/spuco\_animals/spuco\_animals\_spare.py}
        \item Hyperparameters:
        \begin{itemize}
            \item \texttt{erm\_lr} = [1e-3, 1e-4]
            \item \texttt{erm\_weight\_decay} = [1e-3, 1e-4]
            \item \texttt{lr} = [1e-3, 1e-4]
            \item \texttt{weight\_decay} = [1e-1, 1e-2]
            \item \texttt{infer\_num\_epochs} = [1, 2]
            \item \texttt{num\_clusters} = [2, 4]
            \item \texttt{high\_sampling\_power} = [1, 2]
        \end{itemize}
    \end{itemize}

    \item \textbf{PDE:}
    \begin{itemize}
        \item Command: \texttt{python quickstart/spuco\_animals/spuco\_animals\_pde.py}
        \item Hyperparameters:
        \begin{itemize}
            \item \texttt{lr} = [1e-2, 1e-3, 1e-4]
            \item \texttt{weight\_decay} = [1e-3, 1e-4]
            \item \texttt{warmup\_epochs} = [15, 20]
            \item \texttt{expansion\_size} = [10, 50]
            \item \texttt{expansion\_interval} = [2, 10]
        \end{itemize}
    \end{itemize}

    \item \textbf{JTT:}
    \begin{itemize}
        \item Command: \texttt{python spuco\_animals\_jtt.py}
        \item Hyperparameters:
        \begin{itemize}
            \item \texttt{lr} = [1e-3, 1e-4, 1e-5]
            \item \texttt{weight\_decay} = [1e-5, 1e-4, 1e-3, 1e-2]
            \item \texttt{num\_epochs} = [40, 100]
            \item \texttt{infer\_num\_epochs} = [1, 2]
        \end{itemize}
    \end{itemize}

    \item \textbf{DFR:}
    \begin{itemize}
        \item Command: \texttt{python spuco\_animals\_dfr.py}
        \item Hyperparameters:
        \begin{itemize}
            \item \texttt{lr} = [1e-3, 1e-4, 1e-5]
            \item \texttt{weight\_decay} = [1e-5, 1e-4, 1e-3, 1e-2]
            \item \texttt{num\_epochs} = [40, 100]
        \end{itemize}
    \end{itemize}

    \item \textbf{SSA:}
    \begin{itemize}
        \item Command: \texttt{python spuco\_animals\_ssa.py}
        \item Hyperparameters:
        \begin{itemize}
            \item \texttt{infer\_lr} = [1e-2, 1e-3, 1e-4]
            \item \texttt{infer\_weight\_decay} = [1e-5, 1e-4, 1e-3, 1e-2]
            \item \texttt{infer\_num\_iters} = [1000, 45000]
            \item \texttt{lr} = [1e-3, 1e-4, 1e-5]
            \item \texttt{weight\_decay} = [1e-5, 1e-4, 1e-3, 1e-2]
            \item \texttt{num\_epochs} = [40, 100]
        \end{itemize}
    \end{itemize}
\end{itemize}
\clearpage

\subsection{Additional Results on the Correlation between Group Inference Performance and Worst-group Accuracy}

In this section, we evaluate the quality of inferred groups on the \sun (EASY), \sun (HARD), and \cars datasets. A key distinction between these datasets and \animals is that the former have a group-balanced validation set, while the validation set for \animals mirrors the training set, containing a large majority group with a spurious feature. This difference in validation set distribution significantly increases the size of the minority group (examples without the class's spurious feature) in the former datasets. As a result, group accuracy emerges as a more effective metric than minimum group precision for these datasets. Nonetheless, evaluating group inference on the validation set eliminates the need to train a model with inferred groups solely for tuning hyperparameters during the group inference stage of various methods. Thus, we continue to observe the significant cost savings due to the more efficient model selection offered by evaluation of inferred groups on the validation set. 

\begin{figure*}[h]
    \centering
    \includegraphics[width=.9\textwidth]{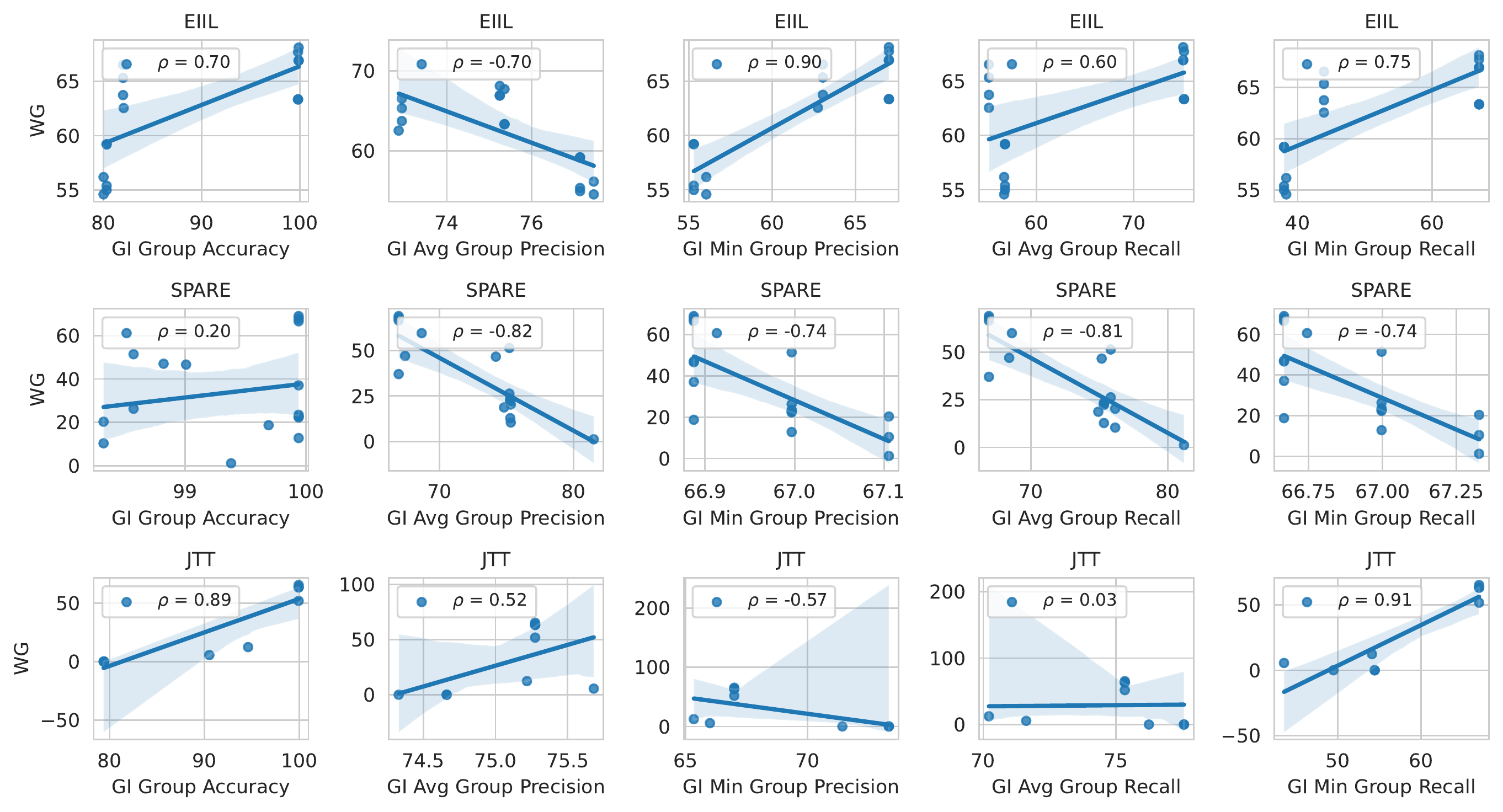}
    \caption{\sun Easy}
    \label{fig:}
\end{figure*}

\begin{figure*}[h]
    \centering
    \includegraphics[width=.9\textwidth]{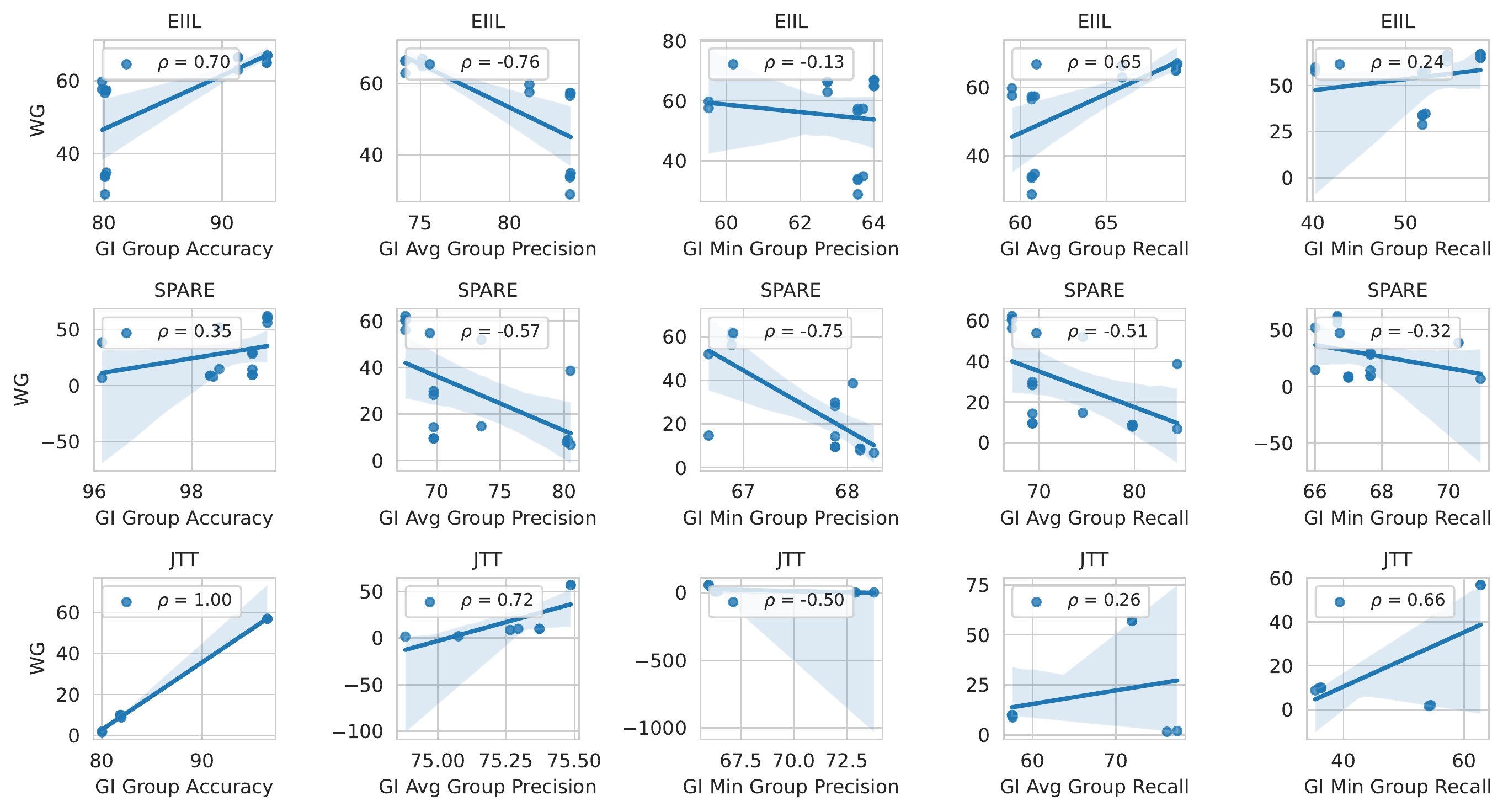}
    \caption{\sun Hard}
    \label{fig:enter-label}
\end{figure*}

\begin{figure*}[h]
    \centering
    \includegraphics[width=.9\textwidth]{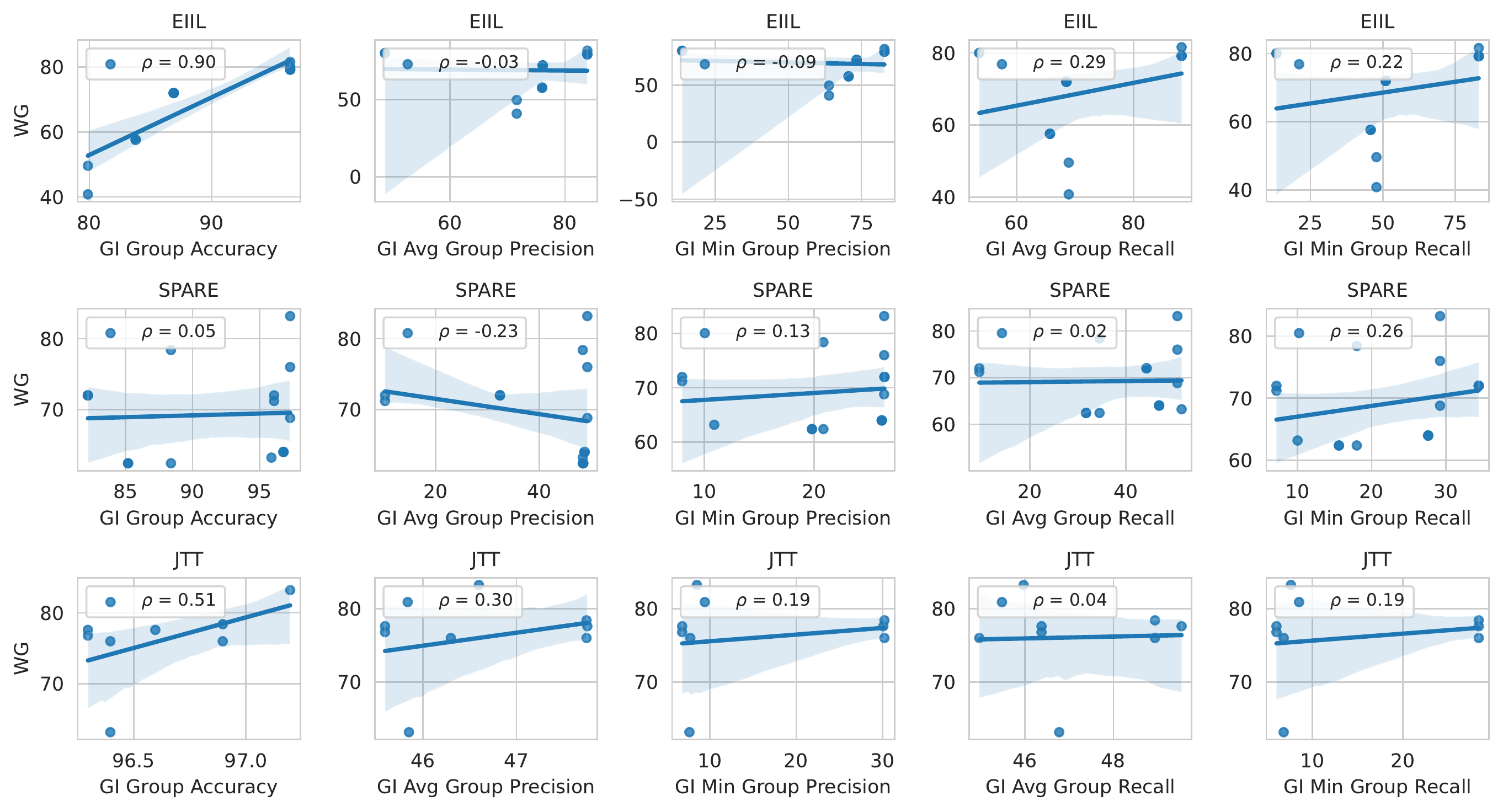}
    \caption{UrbanCars}
    \label{fig:enter-label}
\end{figure*}





\newpage
\section{Additional Results: Pretrained Model Architecture}

Here, we conduct an additional ablation study to assess the impact of architecture on the effectiveness of methods designed to mitigate spurious correlations. Our findings indicate that across all methods, including ERM, a Vision Transformer (ViT) pre-trained with CLIP consistently outperforms a ResNet-50 pre-trained with CLIP. This suggests that the increased number of parameters and/or the architecture of ViT contribute to enhanced robustness against spurious correlations.

\begin{table}[h!]
\centering
\caption{Performances of different methods with ViT.}
\begin{tabular}{lcccc}
\toprule
 & {\textsc{SpuCoSun} (Hard)} & & {\textsc{SpuCoSun} (Easy)} \\
\cmidrule(r){2-3} \cmidrule(r){4-5}
 & WG & Average & WG & Average \\
\midrule
ERM & 47.8 & 97.9 & 38.6 & 97.8 \\\midrule\midrule
SPARE & 86.1\,$\pm$\,0.6 & 95.5\,$\pm$\,0.5 & 84.7\,$\pm$\,0.2 & 97.2\,$\pm$\,0.3 \\
EIIL & 83.1\,$\pm$\,0.1 & 90.9\,$\pm$\,0.1 & 86.7\,$\pm$\,0.1 & 94.7\,$\pm$\,0.1 \\\midrule\midrule
DFR & 87.9\,$\pm$\,0.3 & 94.3\,$\pm$\,0.1 & 89.2\,$\pm$\,0.1 & 92.9\,$\pm$\,0.1 \\\midrule\midrule
GB & 77.7\,$\pm$\,1.0 & 90.6\,$\pm$\,0.5 & 78.9\,$\pm$\,0.6 & 90.5\,$\pm$\,0.1 \\
PDE & 89.2\,$\pm$\,0.2 & 94.0\,$\pm$\,0.5 & 87.5\,$\pm$\,1.0 & 94.5\,$\pm$\,0.1 \\

\bottomrule
\end{tabular}
\label{tab:architecture_compare}
\end{table}

\end{document}